\documentclass{article} 
\usepackage{iclr2027_conference,times}

\usepackage{amsmath,amsfonts,bm}

\def\eqref#1{equation~\ref{#1}}

\def\1{\bm{1}}

\DeclareMathAlphabet{\mathsfit}{\encodingdefault}{\sfdefault}{m}{sl}
\SetMathAlphabet{\mathsfit}{bold}{\encodingdefault}{\sfdefault}{bx}{n}

\usepackage{hyperref}
\usepackage{url}
\usepackage{graphicx}
\usepackage{booktabs}
\usepackage{multirow}
\usepackage{algorithm}
\usepackage{algorithmic}
\usepackage{subcaption}

\newcommand{\ours}{Attention-Patching}

\title{Learn the Effects, Not the Examples: Implicit ICL via Attention Patching}

\author{%
Seungwon Woo\thanks{Equal contribution.},
Hyunseo Shin\footnotemark[1],
Wonseok Hwang\thanks{Corresponding author.} \\
University of Seoul \\
\texttt{\{wsw20047,hseo98,wonseok.hwang\}@uos.ac.kr}
}

\iclrfinalcopy 
\begin{document}

\maketitle
\lhead{} 

\begin{abstract}
In-context learning (ICL) enables language models to adapt to new tasks from demonstrations, but repeatedly including multiple examples increases context length and inference cost. Many existing implicit ICL methods encode demonstrations as fixed vectors, limiting adaptation of the intervention to individual queries and token positions. We propose \ours, an implicit ICL method that models contextual changes in self-attention outputs through task-specific updates to attention output projections. Using query-answer pairs processed with and without a fixed demonstration set, \ours\ fits layer-wise affine transformations from non-contextual attention representations to the demonstration-induced output changes. Across 12 classification benchmarks and four model families, \ours\ achieves higher average accuracy than the evaluated implicit ICL baselines while maintaining performance comparable to explicit ICL at near zero-shot inference cost. Further experiments on diverse generation tasks demonstrate its applicability to autoregressive multi-token generation.

\end{abstract}

\section{Introduction}
In-context learning (ICL) enables large language models (LLMs) to adapt to new tasks from a small set of input-output demonstrations without gradient-based parameter updates \cite{brown2020language,wies2023the, schoch-ji-2025-good,min-etal-2022-rethinking}. In explicit ICL, however, the demonstrations must be included in the prompt for every query \cite{hao2022structured,li2025longcontext}. This increases sequence length and the associated prefill computation and memory requirements, particularly when the same task is repeatedly applied to many queries. 

Recent studies address this cost through implicit ICL, which encodes demonstrations into task representations and applies them during inference without retaining the demonstration tokens. These representations are commonly derived from contextual hidden states, attention-head outputs, or activation differences \cite{hendel-etal-2023-context,todd2024function, liu2024incontext,liu2025iterative,li2025implicit}. Prior results show that such representations can recover a substantial portion of the task behavior induced by explicit demonstrations while using shorter inference prompts.

While vector-based implicit ICL avoids retaining demonstrations in the inference prompt, it typically represents their effect using a fixed vector.
Such a representation cannot account for the fact that the influence of the same demonstrations may vary with the current query and token representation.

To address this limitation, we propose \textbf{\ours}, which learns task-specific transformations that map attention representations to the activation shifts observed under explicit ICL. By fitting layer-wise affine mappings over paired forward passes, our method enables adaptive steering, producing input-dependent corrections in both direction and magnitude rather than applying a fixed steering direction. These mappings are incorporated directly into the attention output projections, preserving the original inference architecture without introducing additional matrix multiplications.


We evaluate \ours\ on 12 classification benchmarks across four model families. The results demonstrate consistent improvements in average accuracy over the evaluated implicit ICL baselines, with performance approaching or exceeding explicit few-shot ICL depending on the model. Beyond classification, experiments on diverse generation tasks provide evidence that the learned patches extend to autoregressive multi-token generation. Because the learned patches are merged into the model weights, per-query inference cost of \ours\ remains close to zero-shot inference even when the task representation is derived from long demonstration contexts.

Overall, our contributions are threefold:
\begin{itemize}
    \item We introduce \textbf{\ours}, which represents the contextual effect of demonstrations as task-specific updates to the attention output projections.
    
    \item Across classification and multi-token generation tasks, \ours\ generally outperforms evaluated implicit ICL baselines while remaining competitive with explicit ICL and maintaining inference cost close to zero-shot inference.

    \item We analyze the adaptive steering behavior of the learned patches, showing that input-dependent weight transformations account for most of their performance and characterizing their structure across singular directions and model layers.
    
\end{itemize}

\section{Related Work}

\subsection{Implicit In-Context Learning}
Implicit ICL seeks to encode demonstrations into internal task representations that can be reused during query-only inference. Early approaches primarily differ in the activation source used to construct these representations. \citet{hendel-etal-2023-context} introduce Task
Vectors (TV), which are extracted from hidden states at selected
layers; \citet{todd2024function} construct Function Vectors (FV) by
aggregating the outputs of attention heads associated with ICL, \citet{liu2024incontext} derive In-Context Vectors (ICV) as steering directions from activation differences between paired demonstration inputs and outputs.

More recent work has focused on improving the robustness and expressiveness of these representations, as well as their injection strategies. \citet{zhang2024batch} and \citet{liu2025iterative} aggregate or iteratively refine activation changes across demonstrations, whereas \citet{li2025implicit} integrate attention and MLP representations through calibrated layer-wise coefficients. LTV \cite{saglam-etal-2025-learning} and DyVec \cite{cai-etal-2025-beyond} further replace heuristic choices with learned strategies for composing attention-derived representations and determining their injection locations.

Beyond activation-vector methods, \cite{dherin2025learning} and \cite{mazzawi2026transmutingpromptsweights} propose Thought Patch, showing that contextual effects as implicit updates to downstream MLP weights and study static approximations in controlled settings. \ours\ instead targets the attention output projection, where demonstration-induced changes can be modeled before being propagated through downstream nonlinear dynamics.


\subsection{Attention-based task adaptation}
Prior work has identified specialized attention heads that support different aspects of ICL, including heads that encode task functions, retrieve contextual information, and access parametric knowledge \citep{olsson2022context, yin2025which,kahardipraja2025the}. Complementary theoretical studies in controlled regression and categorical prediction settings show that multi-head attention performs layer-dependent computations that can be interpreted as preprocessing and optimization-like inference \cite{von2023transformers, chen2024how,wang2025on}.

Building on these findings, SITE \citep{park-etal-2026-soft} learns task-specific head-wise coefficients to interpolate current head activations with fixed task embeddings extracted from few-shot prompts. In-Context Routing (ICR) \citep{li2026train} trains a query-conditioned router on multiple tasks to combine cross-task query-key directions and modify attention logits.

These findings further motivate us to encode demonstration effects at the attention output projection, where information across attention heads is combined. Unlike SITE and ICR, \ours\ fits task-specific affine patches via ridge regression on activations collected from paired forward passes, without backpropagation through the language model.
\section{Method}

\begin{figure}[!h]
\centering
\includegraphics[width=1\linewidth]{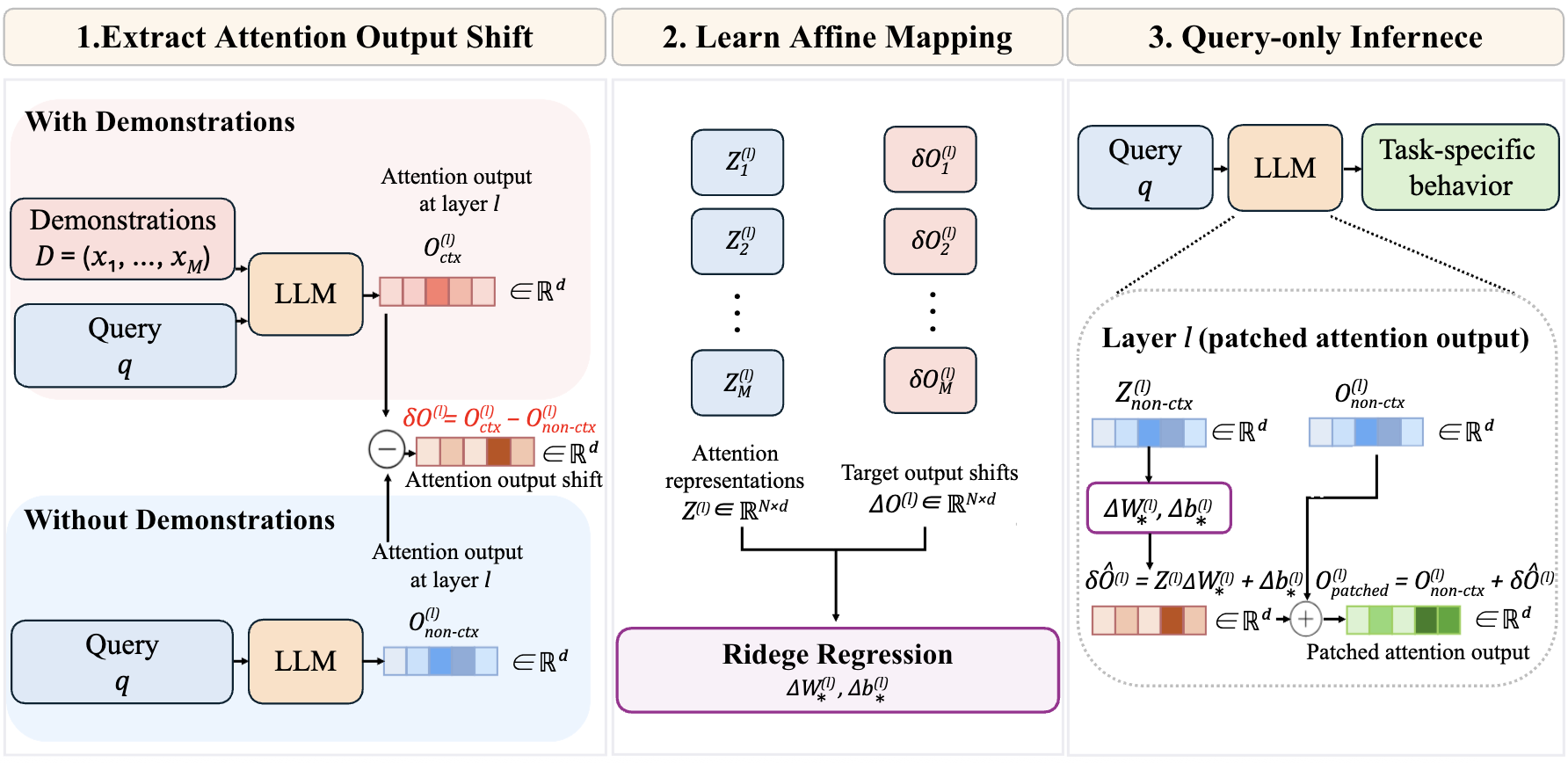}
\caption{Framework of \ours. Attention output shifts induced by demonstrations are used to learn task-specific mappings, incorporated into the attention output projections for inference.}
\label{fig:framework}
\end{figure}

In this section, we introduce \ours\ by defining demonstration effect in attention (Section~\ref{sec:demo_effect}) and constructing affine attention patches to capture these effects (Section~\ref{sec:attn_patching}). We then provide an adaptive steering view of the learned patches in Section~\ref{sec:adap_steering}.


\subsection{Demonstration Effect in Attention}
\label{sec:demo_effect}

Let \(\mathcal{D}\) denote an ordered sequence of demonstrations, and let \(\mathcal Q=\{(x_i,y_i)\}_{i=1}^{M}\) denote \(M\) query–answer pairs. For a query-answer pair \((x,y) \in  \mathcal{Q}\), we consider two prompts that contain the same query-answer pair but differ in whether the demonstrations are included:
\[
p_{\mathrm{ctx}}=[\mathcal{D};x;y],
\qquad
p_{\mathrm{non\text{-}ctx}}=[x;y].
\]

We compare the attention outputs at the token positions corresponding to the query-answer pair under these two prompts. At layer \(\ell\), let
\[
O_{\mathrm{ctx}}^{(\ell)}(x,y),
\;
O_{\mathrm{non\text{-}ctx}}^{(\ell)}(x,y)
\in \mathbb{R}^{n\times d_{\mathrm{out}}}
\]
denote the corresponding attention outputs. Here, \(n\) is the number of tokens in \((x,y)\), and \(d_{\mathrm{out}}\) is the representation dimension. We define the layer-wise demonstration effect as
\[
\delta O^{(\ell)}(x,y)
=
O_{\mathrm{ctx}}^{(\ell)}(x,y)
-
O_{\mathrm{non\text{-}ctx}}^{(\ell)}(x,y).
\]

\subsection{\ours}
\label{sec:attn_patching}

\paragraph{Calibration}

At each layer \(\ell\), the attention output is obtained by applying the output projection \((W_O^{(\ell)}, b_O^{(\ell)})\) to the attention representation \(Z^{(\ell)}\): 
\[
O^{(\ell)} 
= 
Z^{(\ell)}W_O^{(\ell)} 
+ 
\mathbf{1}_n b_O^{(\ell)\top}, 
\qquad 
Z^{(\ell)} 
= 
\operatorname{Concat}_{h=1}^{H} \left(z_h^{(\ell)}\right),
\]
where \(z_h^{(\ell)}\) denotes the output of attention head \(h\). Accordingly, the demonstration effect defined in Section~\ref{sec:demo_effect} can be expressed as 
\[ 
\delta O^{(\ell)}(x,y) 
= 
\left( Z_{\mathrm{ctx}}^{(\ell)}(x,y) 
- 
Z_{\mathrm{non\text{-}ctx}}^{(\ell)}(x,y) \right) W_O^{(\ell)}, 
\] 
where the shared output-projection bias cancels between the two forward passes.

Motivated by the affine structure of the output projection, we model the demonstration effect from the non-contextual attention representation as 
\[ 
\widehat{\delta O}^{(\ell)}(x,y) 
= 
Z_{\mathrm{non\text{-}ctx}}^{(\ell)}(x,y) \Delta W^{(\ell)} 
+ 
\mathbf{1}_n \Delta b^{(\ell)\top}, 
\] 
where \((\Delta W^{(\ell)}, \Delta b^{(\ell)})\) parameterize the layer-wise attention patch. 

Combining the demonstration effect with the non-contextual attention output gives
\[
\begin{aligned}
O_{\mathrm{ctx}}^{(\ell)}
&=
O_{\mathrm{non\text{-}ctx}}^{(\ell)}
+
\delta O^{(\ell)} \\
&=
Z_{\mathrm{non\text{-}ctx}}^{(\ell)}W_O^{(\ell)}
+
\mathbf{1}_n b_O^{(\ell)\top}
+
\left(
Z_{\mathrm{ctx}}^{(\ell)}
-
Z_{\mathrm{non\text{-}ctx}}^{(\ell)}
\right)W_O^{(\ell)} \\
&\approx
Z_{\mathrm{non\text{-}ctx}}^{(\ell)}
\left(
W_O^{(\ell)}+\Delta W^{(\ell)}
\right)
+
\mathbf{1}_n
\left(
b_O^{(\ell)}+\Delta b^{(\ell)}
\right)^\top.
\end{aligned}
\]
Thus, approximating the demonstration effect with the affine patch is equivalent to approximating the contextual attention output through an update to the original output projection.

To calibrate the patch parameters, we collect the non-contextual attention representations and their corresponding demonstration effects over the calibration set \(\mathcal{Q}\). Let \(\mathbf{Z}^{(\ell)} \in \mathbb{R}^{N\times d_{\mathrm{in}}}\) and \(\mathbf{\Delta O}^{(\ell)} \in \mathbb{R}^{N\times d_{\mathrm{out}}}\) denote the row-wise concatenations of \(Z_{\mathrm{non\text{-}ctx}}^{(\ell)}(x,y)\) and \(\delta O^{(\ell)}(x,y)\) over all \((x,y)\in\mathcal Q\), respectively, where \(N\) is the total number of query-answer tokens in the calibration set. 

The layer-wise attention patch is then fitted by ridge regression:
\[
\left(
\Delta W_*^{(\ell)},
\Delta b_*^{(\ell)}
\right)
=
\arg\min_{\Delta W,\Delta b}
\frac{1}{N}
\left\|
\mathbf{Z}^{(\ell)}\Delta W
+
\mathbf{1}_N\Delta b^\top
-
\mathbf{\Delta O}^{(\ell)}
\right\|_F^2
+
\rho\|\Delta W\|_F^2,
\]
where \(\rho>0\) controls the regularization strength. 
This objective admits the following closed-form solution:
\[
\Delta W_*^{(\ell)}
=
\left(
\bar{\mathbf{Z}}^{(\ell)\top}
\bar{\mathbf{Z}}^{(\ell)}
+
N\rho I
\right)^{-1}
\bar{\mathbf{Z}}^{(\ell)\top}
\overline{\mathbf{\Delta O}}^{(\ell)},
\qquad
\Delta b_*^{(\ell)\top}
=
\mu_{\Delta O}^{(\ell)}
-
\mu_Z^{(\ell)}\Delta W_*^{(\ell)},
\]
where \(\bar{\mathbf{Z}}^{(\ell)}\) and \(\overline{\mathbf{\Delta O}}^{(\ell)}\) denote the centered versions of \(Z^{(\ell)}\) and \(\Delta O^{(\ell)}\), respectively. The full derivation and definitions are provided in Appendix~\ref{app:closed_form}, with a functional view of the learned patches presented in Appendix~\ref{app:func_view}.

\paragraph{Inference}

To reduce sensitivity to the calibration set \(\mathcal{Q}\), we introduce a scalar patch strength \(\alpha\) that controls the magnitude of the calibrated patch. After calibration, the scaled layer-wise patches are merged into the original attention output projections:
\[
W_O^{(\ell)}
\leftarrow
W_O^{(\ell)}+\alpha \Delta W_\star^{(\ell)},
\qquad
b_O^{(\ell)}
\leftarrow
b_O^{(\ell)}+\alpha \Delta b_\star^{(\ell)}.
\]
The coefficient $\alpha$ therefore controls the strength of the learned contextual correction while preserving the same parameterization as the original attention output projection. Once merged, the patch is applied through the existing projection and requires no additional matrix multiplication or intervention module at inference time. The overall framework is depicted in Figure~\ref{fig:framework}, and the pseudocode is provided in Algorithm~\ref{alg:patching} (Appendix~\ref{app:pseudocode}).

\subsection{Adaptive Steering with \ours}
\label{sec:adap_steering}

At token position \(t\), applying the calibrated attention patch with strength \(\alpha\) yields
\[
\widehat{\delta O^t}
=
\alpha\left(
Z_{\mathrm{non\text{-}ctx}}^{t}\Delta W_\star
+
\Delta b_\star^\top
\right),
\]
where we omit the layer index for clarity. Using the singular value decomposition
\(\Delta W_\star = U\Sigma V^\top = \sum_i \sigma_i u_i v_i^\top\),
the correction becomes
\[
\widehat{\delta O^t}
=
\alpha\left(
\sum_i
\sigma_i
\bigl(Z_{\mathrm{non\text{-}ctx}}^{t}u_i\bigr)
v_i^\top
+
\Delta b_\star^\top
\right).
\]
This decomposition provides an adaptive steering view of \ours. The right singular vectors \(v_i\) define task-specific steering directions, with coefficients \(\sigma_i(Z_{\mathrm{non\text{-}ctx}}^t u_i)\) that vary with the current attention representation across inputs and token positions. The bias term \(\Delta b_\star^\top\) provides a shared additive correction, and the patch strength \(\alpha\) controls the overall correction magnitude.

\section{Experiments}

\subsection{Experimental Setup}
\label{sec:experimental_setup}
\paragraph{Models} 
We evaluate \ours\ (AP) using four open-weight language model families: Llama-3.2-3B-Instruct \citep{grattafiori2024llama}, Qwen3.5-0.8B \citep{qwen3.5}, Mistral-7B-Instruct-v0.3 \citep{jiang2023mistral7b}, Gemma-2-27b-it and  Gemma-2-2b-it \citep{gemmateam2024gemma2improvingopen}. 

\paragraph{Baselines} 
Across 12 classification benchmarks, we compare \ours\ with 0-shot and explicit few-shot ICL. We also include LoRA~\citep{hu2022lora} and SITE~\citep{park-etal-2026-soft}, both of which require gradient-based optimization. We consider six vector-based implicit ICL baselines: Task Vector (TV)~\citep{hendel-etal-2023-context}, Function Vector (FV)~\citep{todd2024function}, Iterative Vectors (IV)~\citep{liu2025iterative}, Implicit In-context Learning (I2CL)~\citep{li2025implicit}, ICV~\citep{liu2024incontext}, and Linear Task Vector (LTV)~\citep{kwon2026distributional}. We also compare against Thought Patching (TP)~\citep{mazzawi2026transmutingpromptsweights}. For all baselines, we follow the experimental details reported in their original papers.

\begin{table*}[!t]
\caption{
Baseline comparison across classification benchmarks.
All results are averaged over five independent runs.
The best and second-best results in each column within each model
are indicated in bold and underline, respectively. Method abbreviations are defined in Section~\ref{sec:experimental_setup}.
}
\label{tab:classification}
\centering
\resizebox{\textwidth}{!}{%
\begin{tabular}{@{}lc|*{12}{c}@{}}

 \toprule
Method & Avg. & AG & SST-2 & TREC & CSQA & PIQA & SST-5 & MR & MRPC & CB & COPA & CREAK & AI2SciE \\
\midrule
\multicolumn{14}{c}{\textbf{Llama-3.2-3B-Instruct}} \\
\midrule
0-shot 
& 73.5 & 58.8 & 90.8 & 62.0 & \textbf{76.6} & 75.6 & 33.8 & 89.0 & 66.9 & 67.9 & \underline{94.0} & 79.2 & 87.0 
\\

ICL 
& \underline{78.6} & 86.4{\tiny ($\pm$1.5)} & \textbf{93.5}{\tiny ($\pm$0.5)} & 79.9{\tiny ($\pm$4.5)} & 73.4{\tiny ($\pm$1.0)} & 74.9{\tiny ($\pm$2.0)} & \textbf{47.8}{\tiny ($\pm$2.6)} & \textbf{91.2}{\tiny ($\pm$0.9)} & 50.7{\tiny ($\pm$7.1)} & \textbf{83.6}{\tiny ($\pm$4.6)} & 88.8{\tiny ($\pm$1.9)} & \textbf{86.0}{\tiny ($\pm$0.4)} & 86.9{\tiny ($\pm$0.7)} 
\\

\midrule
LoRA 
& 77.1 & \underline{87.4}{\tiny ($\pm$0.8)} & 91.1{\tiny ($\pm$0.6)} & \textbf{86.3}{\tiny ($\pm$1.7)} & \underline{76.3}{\tiny ($\pm$0.5)} & 76.0{\tiny ($\pm$0.5)} & 43.8{\tiny ($\pm$1.8)} & 88.8{\tiny ($\pm$0.8)} & 56.3{\tiny ($\pm$5.2)} & 64.6{\tiny ($\pm$1.5)} & \textbf{94.4}{\tiny ($\pm$0.5)} & 73.0{\tiny ($\pm$2.9)} & 87.2{\tiny ($\pm$0.2)} 
\\

SITE 
& 77.9 & \textbf{88.8}{\tiny ($\pm$1.7)} & 91.7{\tiny ($\pm$0.5)} & 78.3{\tiny ($\pm$8.4)} & 73.4{\tiny ($\pm$0.7)} & 75.8{\tiny ($\pm$2.5)} & 38.2{\tiny ($\pm$1.7)} & 90.4{\tiny ($\pm$0.4)} & \underline{68.4}{\tiny ($\pm$0.0)} & 67.9{\tiny ($\pm$10.3)} & 91.0{\tiny ($\pm$4.3)} & 84.1{\tiny ($\pm$1.3)} & 87.0{\tiny ($\pm$1.4)} 
\\

\midrule

TV 
& 71.3 & 68.8{\tiny ($\pm$3.8)} & 90.7{\tiny ($\pm$0.3)} & 69.9{\tiny ($\pm$2.7)} & 76.2{\tiny ($\pm$0.2)} & \underline{76.1}{\tiny ($\pm$1.1)} & 31.0{\tiny ($\pm$5.4)} & 90.2{\tiny ($\pm$0.6)} & 40.9{\tiny ($\pm$13.1)} & 54.3{\tiny ($\pm$19.1)} & 90.8{\tiny ($\pm$2.4)} & 79.4{\tiny ($\pm$1.2)} & 87.4{\tiny ($\pm$0.3)} 
\\

I2CL 
& 74.6 & 83.3{\tiny ($\pm$3.3)} & 86.3{\tiny ($\pm$2.6)} & 79.5{\tiny ($\pm$7.3)} & 68.8{\tiny ($\pm$2.2)} & 74.2{\tiny ($\pm$2.0)} & 41.9{\tiny ($\pm$2.6)} & 88.4{\tiny ($\pm$1.9)} & 59.4{\tiny ($\pm$4.7)} & 65.0{\tiny ($\pm$6.9)} & 90.2{\tiny ($\pm$3.4)} & 72.8{\tiny ($\pm$4.2)} & 85.1{\tiny ($\pm$1.4)} 
\\

FV 
& 73.5 & 61.1{\tiny ($\pm$1.2)} & 91.4{\tiny ($\pm$0.6)} & 81.4{\tiny ($\pm$0.8)} & 71.9{\tiny ($\pm$0.9)} & 74.2{\tiny ($\pm$1.6)} & 31.2{\tiny ($\pm$1.2)} & 88.5{\tiny ($\pm$0.4)} & 53.0{\tiny ($\pm$1.5)} & \underline{82.0}{\tiny ($\pm$1.0)} & 82.5{\tiny ($\pm$1.9)} & 77.4{\tiny ($\pm$1.6)} & \textbf{87.8}{\tiny ($\pm$0.5)} 
\\

IV 
& 72.6 & 84.8{\tiny ($\pm$2.1)} & 92.6{\tiny ($\pm$0.4)} & 70.0{\tiny ($\pm$1.9)} & 67.6{\tiny ($\pm$1.0)} & 73.2{\tiny ($\pm$1.3)} & 45.9{\tiny ($\pm$2.1)} & 88.2{\tiny ($\pm$0.4)} & 31.6{\tiny ($\pm$0.0)} & 70.5{\tiny ($\pm$3.7)} & 74.7{\tiny ($\pm$1.3)} & \underline{85.9}{\tiny ($\pm$1.2)} & 86.6{\tiny ($\pm$0.5)} 
\\

ICV 
& 74.5 & 64.4{\tiny ($\pm$0.7)} & 88.5{\tiny ($\pm$1.5)} & 71.0{\tiny ($\pm$2.6)} & 75.8{\tiny ($\pm$0.3)} & \textbf{76.4}{\tiny ($\pm$0.7)} & 35.7{\tiny ($\pm$0.4)} & 88.9{\tiny ($\pm$0.3)} & 66.9{\tiny ($\pm$0.0)} & 67.9{\tiny ($\pm$0.0)} & \underline{94.0}{\tiny ($\pm$0.0)} & 79.0{\tiny ($\pm$0.4)} & 85.9{\tiny ($\pm$0.5)} 
\\

LTV 
& 76.6 & 79.3{\tiny ($\pm$3.1)} & 92.0{\tiny ($\pm$1.2)} & 81.1{\tiny ($\pm$3.6)} & 71.9{\tiny ($\pm$2.2)} & 72.6{\tiny ($\pm$3.8)} & 42.4{\tiny ($\pm$1.7)} & 89.4{\tiny ($\pm$1.5)} & 49.0{\tiny ($\pm$4.3)} & 78.9{\tiny ($\pm$2.9)} & 92.2{\tiny ($\pm$2.7)} & 82.2{\tiny ($\pm$2.3)} & \underline{87.7}{\tiny ($\pm$1.4)} 
\\

\midrule
TP 
& 70.8 & 74.6{\tiny ($\pm$3.6)} & 88.0{\tiny ($\pm$0.9)} & 71.4{\tiny ($\pm$5.2)} & 62.4{\tiny ($\pm$2.0)} & 74.6{\tiny ($\pm$1.2)} & 38.4{\tiny ($\pm$2.9)} & 88.4{\tiny ($\pm$1.2)} & 64.4{\tiny ($\pm$4.7)} & 78.6{\tiny ($\pm$4.4)} & 88.2{\tiny ($\pm$2.2)} & 75.2{\tiny ($\pm$3.0)} & 45.7{\tiny ($\pm$5.1)} 
\\

\textbf{AP (Ours)} 
& \textbf{80.7} & 85.8{\tiny ($\pm$2.9)} & \underline{93.0}{\tiny ($\pm$0.6)} & \underline{83.9}{\tiny ($\pm$1.5)} & 76.2{\tiny ($\pm$0.4)} & 75.8{\tiny ($\pm$0.2)} & \underline{46.7}{\tiny ($\pm$2.5)} & \underline{91.1}{\tiny ($\pm$0.8)} & \textbf{68.7}{\tiny ($\pm$0.8)} & 79.6{\tiny ($\pm$7.2)} & 93.2{\tiny ($\pm$1.3)} & \textbf{86.0}{\tiny ($\pm$1.3)} & \underline{87.7}{\tiny ($\pm$0.8)} 
\\

\midrule

\multicolumn{14}{c}{\textbf{Gemma-2-27b-it}} \\
\midrule
0-shot 
& 82.2 & 78.2 & 95.2 & 77.6 & 80.4 & 88.8 & 39.6 & \textbf{93.8} & 63.7 & 85.7 & 98.0 & 87.6 & \underline{97.8} 
\\

ICL 
& \textbf{87.9} & 88.6{\tiny ($\pm$1.0)} & 94.7{\tiny ($\pm$0.6)} & \underline{91.6}{\tiny ($\pm$2.4)} & \textbf{85.7}{\tiny ($\pm$0.4)} & \textbf{90.4}{\tiny ($\pm$0.5)} & \textbf{55.3}{\tiny ($\pm$2.3)} & \underline{93.7}{\tiny ($\pm$0.2)} & 75.0{\tiny ($\pm$1.6)} & 90.0{\tiny ($\pm$2.0)} & \textbf{100.0}{\tiny ($\pm$0.0)} & \textbf{92.7}{\tiny ($\pm$0.6)} & 97.0{\tiny ($\pm$0.2)} 
\\

\midrule

LoRA 
& 86.9 & 89.4{\tiny ($\pm$0.9)} & 94.6{\tiny ($\pm$1.2)} & \textbf{91.8}{\tiny ($\pm$2.7)} & 82.8{\tiny ($\pm$0.6)} & 87.5{\tiny ($\pm$1.2)} & 52.5{\tiny ($\pm$4.6)} & 91.7{\tiny ($\pm$3.6)} & \textbf{77.0}{\tiny ($\pm$2.7)} & 87.9{\tiny ($\pm$2.6)} & 99.0{\tiny ($\pm$0.7)} & 91.4{\tiny ($\pm$1.4)} & 97.6{\tiny ($\pm$0.3)} 
\\

SITE 
& 85.1 & \textbf{90.3}{\tiny ($\pm$1.2)} & 95.5{\tiny ($\pm$0.5)} & 84.7{\tiny ($\pm$3.4)} & 82.3{\tiny ($\pm$0.6)} & 87.8{\tiny ($\pm$1.3)} & 40.4{\tiny ($\pm$1.5)} & 93.6{\tiny ($\pm$0.5)} & 68.9{\tiny ($\pm$1.1)} & 90.4{\tiny ($\pm$1.0)} & 98.8{\tiny ($\pm$0.4)} & 91.4{\tiny ($\pm$1.2)} & 97.1{\tiny ($\pm$0.2)} 
\\

\midrule

TV 
& 81.5 & 83.1{\tiny ($\pm$2.0)} & 95.0{\tiny ($\pm$0.2)} & 79.0{\tiny ($\pm$1.4)} & 80.8{\tiny ($\pm$0.3)} & 88.1{\tiny ($\pm$0.5)} & 39.2{\tiny ($\pm$0.3)} & 93.6{\tiny ($\pm$0.4)} & 56.5{\tiny ($\pm$8.2)} & 80.7{\tiny ($\pm$0.8)} & 98.0{\tiny ($\pm$0.0)} & 86.0{\tiny ($\pm$1.2)} & 97.5{\tiny ($\pm$0.1)} 
\\

I2CL 
& 82.7 & 84.4{\tiny ($\pm$3.9)} & 92.6{\tiny ($\pm$1.2)} & 82.4{\tiny ($\pm$5.1)} & 80.2{\tiny ($\pm$0.4)} & 87.7{\tiny ($\pm$1.1)} & 42.1{\tiny ($\pm$3.4)} & 92.3{\tiny ($\pm$1.0)} & 64.7{\tiny ($\pm$11.7)} & 83.6{\tiny ($\pm$5.6)} & 99.2{\tiny ($\pm$0.5)} & 85.9{\tiny ($\pm$3.3)} & 97.2{\tiny ($\pm$0.3)}
\\

FV 
& 85.0 & 88.0{\tiny ($\pm$0.6)} & \textbf{96.0}{\tiny ($\pm$0.5)} & 85.5{\tiny ($\pm$0.2)} & 82.6{\tiny ($\pm$0.6)} & 86.4{\tiny ($\pm$0.9)} & 47.4{\tiny ($\pm$1.4)} & 92.9{\tiny ($\pm$0.4)} & 72.4{\tiny ($\pm$0.8)} & 90.4{\tiny ($\pm$0.2)} & 98.8{\tiny ($\pm$0.4)} & 82.0{\tiny ($\pm$0.8)} & 97.4{\tiny ($\pm$0.3)} 
\\

IV 
& 86.3 & \underline{89.5}{\tiny ($\pm$0.6)} & 94.8{\tiny ($\pm$0.2)} & 87.9{\tiny ($\pm$0.8)} & \underline{83.4}{\tiny ($\pm$1.0)} & 85.9{\tiny ($\pm$0.6)} & 49.1{\tiny ($\pm$0.8)} & 92.5{\tiny ($\pm$0.3)} & \underline{75.4}{\tiny ($\pm$1.1)} & \underline{92.4}{\tiny ($\pm$0.7)} & 99.1{\tiny ($\pm$0.1)} & 89.0{\tiny ($\pm$1.0)} & 97.2{\tiny ($\pm$0.4)}
\\

ICV 
& 82.2 & 78.1{\tiny ($\pm$0.2)} & 94.4{\tiny ($\pm$0.5)} & 78.4{\tiny ($\pm$0.1)} & 81.0{\tiny ($\pm$0.4)} & 88.7{\tiny ($\pm$0.4)} & 39.4{\tiny ($\pm$0.1)} & \textbf{93.8}{\tiny ($\pm$0.0)} & 63.5{\tiny ($\pm$0.4)} & 84.6{\tiny ($\pm$1.0)} & 98.0{\tiny ($\pm$0.0)} & 89.3{\tiny ($\pm$1.7)} & 97.6{\tiny ($\pm$0.1)} 
\\

LTV 
& 85.9 & 88.2{\tiny ($\pm$1.0)} & \underline{95.8}{\tiny ($\pm$0.7)} & 88.3{\tiny ($\pm$4.0)} & 81.3{\tiny ($\pm$0.9)} & 86.8{\tiny ($\pm$2.2)} & 50.5{\tiny ($\pm$3.3)} & 92.6{\tiny ($\pm$1.1)} & 68.0{\tiny ($\pm$5.1)} & 91.1{\tiny ($\pm$1.3)} & 97.8{\tiny ($\pm$0.4)} & \underline{92.1}{\tiny ($\pm$0.9)} & \textbf{98.1}{\tiny ($\pm$0.5)} 
\\

\midrule

TP 
& 84.7 & 83.8{\tiny ($\pm$2.0)} & 94.8{\tiny ($\pm$0.7)} & 84.1{\tiny ($\pm$2.1)} & \underline{83.4}{\tiny ($\pm$0.7)} & \underline{88.9}{\tiny ($\pm$0.5)} & 39.6{\tiny ($\pm$0.4)} & \underline{93.7}{\tiny ($\pm$0.3)} & 68.3{\tiny ($\pm$1.6)} & \textbf{93.9}{\tiny ($\pm$1.0)} & 99.0{\tiny ($\pm$0.0)} & 90.0{\tiny ($\pm$0.9)} & 96.5{\tiny ($\pm$0.4)} 
\\

\textbf{AP (Ours)} 
& \underline{87.3} & 89.2{\tiny ($\pm$1.0)} & 95.1{\tiny ($\pm$0.2)} & 91.3{\tiny ($\pm$1.8)} & 81.5{\tiny ($\pm$0.7)} & \underline{88.9}{\tiny ($\pm$0.6)} & \underline{53.8}{\tiny ($\pm$2.2)} & 93.6{\tiny ($\pm$0.3)} & 74.6{\tiny ($\pm$1.2)} & 91.8{\tiny ($\pm$3.7)} & \underline{99.8}{\tiny ($\pm$0.4)} & 90.3{\tiny ($\pm$1.2)} & 97.5{\tiny ($\pm$0.4)} 
\\

\midrule

\multicolumn{14}{c}{\textbf{Qwen3.5-0.8B}} \\
\midrule

0-shot 
& 61.0 & 64.2 & \textbf{88.4} & 62.8 & 49.0 & 59.0 & 28.6 & \textbf{86.4} & \textbf{68.4} & 32.1 & 56.0 & 57.6 & 79.6 
\\

ICL 
& 68.5 & 81.2{\tiny ($\pm$2.4)} & 83.4{\tiny ($\pm$5.2)} & 76.8{\tiny ($\pm$5.9)} & 56.6{\tiny ($\pm$0.7)} & 62.4{\tiny ($\pm$2.1)} & \textbf{43.8}{\tiny ($\pm$2.7)} & \underline{85.4}{\tiny ($\pm$5.2)} & 42.7{\tiny ($\pm$14.2)} & \underline{69.6}{\tiny ($\pm$5.8)} & 62.2{\tiny ($\pm$3.4)} & \underline{72.6}{\tiny ($\pm$3.0)} & \textbf{85.0}{\tiny ($\pm$0.6)} 
\\

\midrule

LoRA 
& 62.8 & 79.4{\tiny ($\pm$3.0)} & 85.9{\tiny ($\pm$1.6)} & 69.5{\tiny ($\pm$7.3)} & 57.1{\tiny ($\pm$0.8)} & 60.5{\tiny ($\pm$1.3)} & 35.0{\tiny ($\pm$1.8)} & 84.4{\tiny ($\pm$1.6)} & 31.9{\tiny ($\pm$0.3)} & 32.9{\tiny ($\pm$4.8)} & 73.4{\tiny ($\pm$1.1)} & 60.0{\tiny ($\pm$3.1)} & 83.5{\tiny ($\pm$0.7)} 
\\

SITE 
& 69.4 & \textbf{84.8}{\tiny ($\pm$2.3)} & 88.2{\tiny ($\pm$1.1)} & 71.6{\tiny ($\pm$14.0)} & 57.2{\tiny ($\pm$1.4)} & 64.1{\tiny ($\pm$4.3)} & 25.1{\tiny ($\pm$4.2)} & \underline{85.4}{\tiny ($\pm$0.8)} & \textbf{68.4}{\tiny ($\pm$0.0)} & 57.1{\tiny ($\pm$13.7)} & \textbf{78.2}{\tiny ($\pm$1.3)} & 69.6{\tiny ($\pm$5.7)} & \underline{83.7}{\tiny ($\pm$1.0)} 
\\

\midrule

TV 
& 66.0 & 73.3{\tiny ($\pm$4.6)} & 86.4{\tiny ($\pm$2.3)} & 64.4{\tiny ($\pm$1.8)} & 53.4{\tiny ($\pm$1.5)} & \textbf{65.2}{\tiny ($\pm$1.0)} & 27.4{\tiny ($\pm$4.0)} & 84.8{\tiny ($\pm$0.4)} & 65.3{\tiny ($\pm$6.9)} & 52.1{\tiny ($\pm$12.7)} & \underline{78.0}{\tiny ($\pm$3.5)} & 58.4{\tiny ($\pm$1.0)} & 83.3{\tiny ($\pm$0.3)} 
\\

I2CL 
& 64.9 & 73.6{\tiny ($\pm$6.5)} & 85.6{\tiny ($\pm$2.9)} & 69.3{\tiny ($\pm$6.8)} & 49.7{\tiny ($\pm$2.9)} & 56.0{\tiny ($\pm$5.2)} & 34.2{\tiny ($\pm$2.2)} & 81.9{\tiny ($\pm$2.7)} & 56.1{\tiny ($\pm$4.1)} & 56.4{\tiny ($\pm$14.7)} & 74.8{\tiny ($\pm$2.6)} & 63.7{\tiny ($\pm$3.1)} & 77.4{\tiny ($\pm$3.6)} 
\\

FV 
& 56.9 & 75.3{\tiny ($\pm$1.5)} & 81.9{\tiny ($\pm$0.9)} & \underline{77.5}{\tiny ($\pm$0.5)} & 26.6{\tiny ($\pm$0.5)} & 53.0{\tiny ($\pm$1.1)} & 33.0{\tiny ($\pm$1.7)} & 76.2{\tiny ($\pm$1.2)} & 33.1{\tiny ($\pm$0.1)} & 52.5{\tiny ($\pm$2.0)} & 45.7{\tiny ($\pm$0.3)} & 62.7{\tiny ($\pm$1.4)} & 64.7{\tiny ($\pm$1.1)} 
\\

IV 
& \underline{70.1} & 80.2{\tiny ($\pm$1.5)} & 83.3{\tiny ($\pm$1.6)} & \textbf{80.9}{\tiny ($\pm$1.0)} & \underline{57.5}{\tiny ($\pm$0.6)} & 62.2{\tiny ($\pm$1.3)} & 37.3{\tiny ($\pm$2.6)} & 82.6{\tiny ($\pm$0.8)} & 68.1{\tiny ($\pm$1.3)} & \textbf{80.0}{\tiny ($\pm$0.5)} & 54.1{\tiny ($\pm$1.9)} & \textbf{72.9}{\tiny ($\pm$1.9)} & 82.3{\tiny ($\pm$0.8)} 
\\

ICV 
& 60.4 & 63.6{\tiny ($\pm$0.3)} & 87.0{\tiny ($\pm$0.8)} & 56.6{\tiny ($\pm$3.2)} & 52.0{\tiny ($\pm$0.8)} & 57.7{\tiny ($\pm$0.5)} & 31.0{\tiny ($\pm$1.4)} & 84.4{\tiny ($\pm$0.6)} & 59.9{\tiny ($\pm$7.3)} & 37.9{\tiny ($\pm$10.4)} & 57.2{\tiny ($\pm$2.9)} & 57.2{\tiny ($\pm$0.3)} & 80.0{\tiny ($\pm$0.4)} 
\\

LTV 
& 64.3 & 76.8{\tiny ($\pm$1.4)} & 82.7{\tiny ($\pm$3.4)} & 66.4{\tiny ($\pm$8.3)} & 53.5{\tiny ($\pm$1.4)} & 59.6{\tiny ($\pm$2.5)} & 32.2{\tiny ($\pm$2.2)} & 84.6{\tiny ($\pm$2.0)} & 56.8{\tiny ($\pm$3.4)} & 48.9{\tiny ($\pm$8.8)} & 67.6{\tiny ($\pm$2.1)} & 61.6{\tiny ($\pm$2.4)} & 81.5{\tiny ($\pm$1.0)} 
\\

\midrule

TP 
& 56.2 & 54.9{\tiny ($\pm$3.4)} & 78.1{\tiny ($\pm$4.6)} & 49.0{\tiny ($\pm$5.2)} & 37.4{\tiny ($\pm$3.9)} & 61.1{\tiny ($\pm$1.2)} & 27.1{\tiny ($\pm$2.4)} & 81.5{\tiny ($\pm$2.3)} & \underline{68.3}{\tiny ($\pm$0.1)} & 58.2{\tiny ($\pm$6.6)} & 57.8{\tiny ($\pm$2.4)} & 59.6{\tiny ($\pm$2.0)} & 41.8{\tiny ($\pm$3.4)} 
\\

\textbf{AP (Ours)} 
& \textbf{70.4} & \underline{81.5}{\tiny ($\pm$1.2)} & \underline{88.3}{\tiny ($\pm$1.5)} & 73.2{\tiny ($\pm$3.8)} & \textbf{59.0}{\tiny ($\pm$2.5)} & \underline{64.2}{\tiny ($\pm$1.9)} & \underline{37.9}{\tiny ($\pm$2.6)} & 85.3{\tiny ($\pm$1.6)} & 67.9{\tiny ($\pm$0.4)} & 63.9{\tiny ($\pm$5.3)} & 73.8{\tiny ($\pm$2.9)} & 66.8{\tiny ($\pm$6.7)} & 83.2{\tiny ($\pm$0.9)} 
\\

\bottomrule
\end{tabular}}
\end{table*}

\paragraph{Evaluation} 
Following \citet{li2026train}, we evaluate each method on up to 500\ instances from the evaluation split of each of the 12 benchmarks, using all available instances for smaller datasets. We evaluate the classification tasks: AGNews \citep{zhang2015character}, SST-2 \citep{socher-etal-2013-recursive}, TREC \citep{li-roth-2002-learning}, CSQA \citep{talmor-etal-2019-commonsenseqa}, PIQA \citep{Bisk2020}, SST-5 \citep{socher-etal-2013-recursive}, MR \citep{pang-lee-2005-seeing}, MRPC \citep{dolan-brockett-2005-automatically}, CB \citep{de-marneffe-etal-2021-universal}, COPA \citep{roemmele_choice_2011}, CREAK \citep{onoe2021creak}, and AI2SciE \citep{allenai:arc}, using five random seeds. For the multi-token prediction task, we evaluate on GSM8K \citep{cobbe2021gsm8k}, FLORES (Korean and Spanish subsets) \citep{goyal2022flores}, DailyMail \citep{see-etal-2017-get} and CommonGen \citep{lin-etal-2020-commongen}. Other details are in Appendix \ref{appendix:reproducibility}.

\subsection{Main Results}
\label{main_result}
Table 1 presents the results on Llama-3.2-3B-Instruct, Gemma-2-27b-it, and Qwen3.5-0.8B. Additional results on Mistral-7B-Instruct-v0.3 and Gemma-2-2b-it are provided in Appendix \ref{appendix:one_token_prediction}.

\paragraph{\ours\ outperforms implicit ICL baselines.}
As shown in Table~\ref{tab:classification}, \ours\ achieves the highest average accuracy among the evaluated implicit ICL methods across all three models. With 25 calibration examples per class, \ours\ obtains average accuracies of 80.7, 87.3, and 70.4 on Llama-3.2-3B-Instruct, Gemma-2-27b-it, and Qwen3.5-0.8B, respectively, outperforming the strongest implicit ICL baseline on each model by 2.8, 1.0, and 0.3 percentage points.

\paragraph{\ours\ remains competitive with explicit ICL.}
Despite processing only query tokens at inference time, \ours\ achieves performance comparable to explicit ICL across the three models. Specifically, \ours\ exceeds explicit ICL by 2.1 and 1.9 percentage points on Llama and Qwen, respectively, while remaining within 0.6 points on Gemma-2-27b-it. Compared with LoRA, \ours\ also achieves higher average accuracy across all three models, with improvements of 3.6, 0.4, and 7.6 percentage points, respectively. These findings suggest that the learned weight-space transformations can match or exceed the predictive performance of explicit prompting and LoRA on the evaluated classification tasks.

\begin{table}[!t]
\caption{Multi-token generation results under different context lengths, averaged over three independent runs. We report flexible-match accuracy on GSM8K, ROUGE-L F1 and COMET scores on FLORES, and ROUGE-L F1 scores on DailyMail and CommonGen. The best and second-best results in each column within each model and context length are indicated in bold and underline, respectively.}
\scriptsize
\label{tab:generation}

\centering

\resizebox{\columnwidth}{!}{%
\begin{tabular}{l c| c cc cc cc}
  \toprule
  \multirow{2}{*}{Method}
  & \multirow{2}{*}{Avg.}
  & \multirow{2}{*}{GSM8K}
  & \multicolumn{2}{c}{FLORES(Ko)}
  & \multicolumn{2}{c}{FLORES(Sp)}
  & \multirow{2}{*}{DailyMail}
  & \multirow{2}{*}{CommonGen} \\
  \cmidrule(lr){4-5}
  \cmidrule(lr){6-7}
  & & & ROUGE-L F1 & COMET & ROUGE-L F1 & COMET & & \\
  \midrule

  \multicolumn{9}{c}{\textbf{Llama-3.2-3B-Instruct}} \\
  \midrule
  
  
  Zero-shot 
  & 42.9 & 51.4{\tiny ($\pm$0.7)} & 4.8{\tiny ($\pm$0.5)} & 47.3{\tiny ($\pm$0.3)} & 50.6{\tiny ($\pm$0.2)} & 83.2{\tiny ($\pm$0.5)} & 16.8{\tiny ($\pm$0.3)} & 45.8{\tiny ($\pm$0.4)}
  \\
  
  LoRA 
  & 47.4 & 56.1{\tiny ($\pm$2.9)} & 11.4{\tiny ($\pm$1.3)} & 62.7{\tiny ($\pm$3.3)} & 51.1{\tiny ($\pm$0.4)} & 84.2{\tiny ($\pm$0.2)} & 18.1{\tiny ($\pm$1.4)} & 48.2{\tiny ($\pm$0.5)}
  \\
  
  \midrule

  Many-shot (1K) 
  & \textbf{55.1} & \underline{69.6}{\tiny ($\pm$3.3)} & \textbf{21.3}{\tiny ($\pm$1.0)} & \textbf{79.3}{\tiny ($\pm$0.5)} & \underline{51.8}{\tiny ($\pm$0.1)} & \textbf{85.0}{\tiny ($\pm$0.4)} & \textbf{22.1}{\tiny ($\pm$0.9)} & \textbf{56.9}{\tiny ($\pm$1.4)}
  \\
  
  SITE 
  & 47.9 & 69.4{\tiny ($\pm$2.6)} & 8.6{\tiny ($\pm$0.7)} & 56.6{\tiny ($\pm$0.7)} & 50.7{\tiny ($\pm$1.1)} & 84.0{\tiny ($\pm$0.4)} & 16.9{\tiny ($\pm$0.3)} & 49.1{\tiny ($\pm$0.2)} 
  \\
  
  ICV 
  & 43.0 & 52.1{\tiny ($\pm$1.4)} & 5.2{\tiny ($\pm$0.3)} & 46.4{\tiny ($\pm$0.8)} & 51.0{\tiny ($\pm$0.2)} & 83.5{\tiny ($\pm$0.3)} & 16.9{\tiny ($\pm$0.2)} & 46.1{\tiny ($\pm$0.6)} 
  \\
  
  LTV 
  & 38.1 & 39.9{\tiny ($\pm$0.6)} & 10.2{\tiny ($\pm$1.1)} & 45.7{\tiny ($\pm$0.4)} & 49.7{\tiny ($\pm$0.8)} & 82.7{\tiny ($\pm$0.9)} & 17.8{\tiny ($\pm$0.8)} & 20.4{\tiny ($\pm$4.3)} 
  \\
  
  \textbf{AP} 
  & \underline{54.6} & \textbf{70.4}{\tiny ($\pm$5.1)} & \underline{20.4}{\tiny ($\pm$0.7)} & \underline{77.6}{\tiny ($\pm$0.4)} & \textbf{51.9}{\tiny ($\pm$0.4)} & \underline{84.6}{\tiny ($\pm$0.3)} & \underline{21.1}{\tiny ($\pm$1.5)} & \underline{56.4}{\tiny ($\pm$1.1)}
  \\
  
  \midrule

  Many-shot (2K) 
  & \textbf{55.4} & \textbf{70.9}{\tiny ($\pm$1.3)} & \textbf{21.0}{\tiny ($\pm$0.7)} & \textbf{77.0}{\tiny ($\pm$5.0)} & \underline{51.8}{\tiny ($\pm$0.1)} & \textbf{85.3}{\tiny ($\pm$0.2)} & \textbf{24.5}{\tiny ($\pm$0.8)} & \textbf{57.1}{\tiny ($\pm$1.1)}
  \\
  
  SITE 
  & 48.3 & 69.4{\tiny ($\pm$1.2)} & 9.6{\tiny ($\pm$0.9)} & 57.7{\tiny ($\pm$1.4)} & 50.9{\tiny ($\pm$0.6)} & 84.1{\tiny ($\pm$0.3)} & 16.9{\tiny ($\pm$0.4)} & 49.4{\tiny ($\pm$0.1)} 
  \\
  
  ICV 
  & 43.1 & 52.8{\tiny ($\pm$0.9)} & 5.2{\tiny ($\pm$0.4)} & 46.3{\tiny ($\pm$0.3)} & 51.1{\tiny ($\pm$0.3)} & 83.6{\tiny ($\pm$0.3)} & 17.0{\tiny ($\pm$0.2)} & 45.9{\tiny ($\pm$0.4)} 
  \\
  
  LTV 
  & 38.2 & 42.1{\tiny ($\pm$3.8)} & 9.8{\tiny ($\pm$2.0)} & 45.6{\tiny ($\pm$1.2)} & 49.1{\tiny ($\pm$1.8)} & 82.2{\tiny ($\pm$1.6)} & 17.5{\tiny ($\pm$0.6)} & 21.3{\tiny ($\pm$6.1)} 
  \\

  \textbf{AP} 
  & \underline{54.5} & \underline{70.1}{\tiny ($\pm$1.5)} & \underline{19.7}{\tiny ($\pm$0.6)} & \underline{76.0}{\tiny ($\pm$2.2)} & \textbf{52.0}{\tiny ($\pm$0.3)} & \underline{84.6}{\tiny ($\pm$0.3)} & \underline{22.0}{\tiny ($\pm$1.2)} & \underline{57.0}{\tiny ($\pm$1.3)}
  \\

  \midrule

  Many-shot (4K) 
  & \textbf{55.8} & \textbf{70.9}{\tiny ($\pm$0.6)} & \textbf{21.5}{\tiny ($\pm$0.9)} & \textbf{79.7}{\tiny ($\pm$0.4)} & \underline{51.6}{\tiny ($\pm$0.3)} & \textbf{85.2}{\tiny ($\pm$0.4)} & \textbf{25.3}{\tiny ($\pm$1.3)} & \textbf{56.2}{\tiny ($\pm$1.4)}
  \\

  SITE 
  & 48.4 & 70.1{\tiny ($\pm$3.3)} & 10.1{\tiny ($\pm$0.2)} & 57.5{\tiny ($\pm$0.4)} & 51.0{\tiny ($\pm$0.8)} & 84.1{\tiny ($\pm$0.3)} & 16.9{\tiny ($\pm$0.4)} & 49.2{\tiny ($\pm$0.7)}
  \\

  ICV 
  & 43.2 & 52.7{\tiny ($\pm$1.9)} & 5.1{\tiny ($\pm$0.6)} & 46.8{\tiny ($\pm$1.0)} & 51.0{\tiny ($\pm$0.2)} & 83.5{\tiny ($\pm$0.2)} & 16.9{\tiny ($\pm$0.2)} & 46.2{\tiny ($\pm$0.4)} 
  \\
  
  LTV 
  & 38.0 & 42.8{\tiny ($\pm$3.9)} & 10.2{\tiny ($\pm$1.5)} & 46.1{\tiny ($\pm$1.1)} & 47.8{\tiny ($\pm$2.1)} & 81.1{\tiny ($\pm$1.8)} & 17.1{\tiny ($\pm$0.3)} & 21.0{\tiny ($\pm$7.4)}
  \\
  
  \textbf{AP} 
  & \underline{54.9} & \underline{70.4}{\tiny ($\pm$2.3)} & \underline{20.6}{\tiny ($\pm$0.7)} & \underline{77.1}{\tiny ($\pm$0.7)} & \textbf{52.1}{\tiny ($\pm$0.3)} & \underline{84.7}{\tiny ($\pm$0.1)} & \underline{23.4}{\tiny ($\pm$2.4)} & \underline{55.9}{\tiny ($\pm$1.2)}
  \\
  
  \midrule

  Many-shot (8K) 
  & \textbf{55.7} & \underline{70.2}{\tiny ($\pm$1.1)} & \textbf{21.0}{\tiny ($\pm$0.5)} & \textbf{79.7}{\tiny ($\pm$0.3)} & \underline{51.5}{\tiny ($\pm$0.3)} & \textbf{85.3}{\tiny ($\pm$0.1)} & \textbf{26.4}{\tiny ($\pm$0.4)} & \textbf{56.0}{\tiny ($\pm$0.6)}
  \\

  SITE 
  & 48.2 & 68.7{\tiny ($\pm$4.3)} & 9.9{\tiny ($\pm$0.5)} & 57.6{\tiny ($\pm$1.4)} & 51.1{\tiny ($\pm$0.5)} & 84.2{\tiny ($\pm$0.2)} & 16.9{\tiny ($\pm$0.4)} & 49.0{\tiny ($\pm$0.6)}
  \\
  
  ICV 
  & 43.0 & 52.0{\tiny ($\pm$0.9)} & 5.3{\tiny ($\pm$0.5)} & 46.4{\tiny ($\pm$0.8)} & 51.0{\tiny ($\pm$0.3)} & 83.5{\tiny ($\pm$0.3)} & 16.9{\tiny ($\pm$0.2)} & 46.2{\tiny ($\pm$0.7)}
  \\
  
  LTV 
  & 37.4 & 41.5{\tiny ($\pm$3.2)} & 8.9{\tiny ($\pm$0.4)} & 45.7{\tiny ($\pm$0.8)} & 48.0{\tiny ($\pm$1.5)} & 81.2{\tiny ($\pm$1.5)} & 16.9{\tiny ($\pm$0.2)} & 19.2{\tiny ($\pm$5.6)}
  \\
  
  \textbf{AP} 
  & \underline{54.3} & \textbf{70.7}{\tiny ($\pm$4.6)} & \underline{19.0}{\tiny ($\pm$0.7)} & \underline{75.4}{\tiny ($\pm$1.8)} & \textbf{52.0}{\tiny ($\pm$0.4)} & \underline{84.6}{\tiny ($\pm$0.4)} & \underline{23.5}{\tiny ($\pm$0.2)} & \underline{55.1}{\tiny ($\pm$0.6)}
  \\

  
  \midrule

  Many-shot (16K) 
  & \textbf{55.7} & \underline{69.7}{\tiny ($\pm$2.3)} & \textbf{20.8}{\tiny ($\pm$0.9)} & \textbf{79.7}{\tiny ($\pm$0.6)} & \underline{51.6}{\tiny ($\pm$0.2)} & \textbf{85.3}{\tiny ($\pm$0.1)} & \textbf{26.2}{\tiny ($\pm$0.4)} & \textbf{56.9}{\tiny ($\pm$1.4)}
  \\

  SITE 
  & 47.9 & 69.4{\tiny ($\pm$1.2)} & 8.7{\tiny ($\pm$0.3)} & 56.0{\tiny ($\pm$1.5)} & 50.7{\tiny ($\pm$0.8)} & 83.9{\tiny ($\pm$0.5)} & 17.0{\tiny ($\pm$0.5)} & 49.3{\tiny ($\pm$0.6)}
  \\

  ICV 
  & 43.0 & 51.8{\tiny ($\pm$1.8)} & 5.2{\tiny ($\pm$0.3)} & 46.4{\tiny ($\pm$0.9)} & 50.9{\tiny ($\pm$0.2)} & 83.4{\tiny ($\pm$0.2)} & 17.0{\tiny ($\pm$0.2)} & 46.2{\tiny ($\pm$0.4)} 
  \\
  
  LTV 
  & 36.9 & 41.0{\tiny ($\pm$0.3)} & 8.4{\tiny ($\pm$3.1)} & 44.8{\tiny ($\pm$2.6)} & 46.7{\tiny ($\pm$2.4)} & 80.3{\tiny ($\pm$2.2)} & 16.8{\tiny ($\pm$0.4)} & 20.0{\tiny ($\pm$7.6)}
  \\
  
  \textbf{AP} 
  & \underline{54.6} & \textbf{70.0}{\tiny ($\pm$0.0)} & \underline{20.0}{\tiny ($\pm$1.2)} & \underline{76.3}{\tiny ($\pm$1.2)} & \textbf{52.0}{\tiny ($\pm$0.2)} & \underline{84.7}{\tiny ($\pm$0.1)} & \underline{24.1}{\tiny ($\pm$0.3)} & \underline{55.1}{\tiny ($\pm$1.1)}
  \\
  

  \midrule
  \multicolumn{9}{c}{\textbf{Gemma-2-27b-it}} \\
  \midrule
  Zero-shot 
  & 58.7 & 79.8{\tiny ($\pm$0.5)} & 23.5{\tiny ($\pm$0.8)} & 89.4{\tiny ($\pm$0.2)} & 55.7{\tiny ($\pm$0.1)} & 86.9{\tiny ($\pm$0.2)} & 21.0{\tiny ($\pm$0.3)} & 54.8{\tiny ($\pm$0.4)}
  \\
  
  LoRA 
  & 55.0 & 58.8{\tiny ($\pm$7.9)} & 23.6{\tiny ($\pm$1.0)} & 89.6{\tiny ($\pm$0.1)} & 52.6{\tiny ($\pm$1.7)} & 86.5{\tiny ($\pm$0.8)} & 14.4{\tiny ($\pm$3.4)} & 59.3{\tiny ($\pm$0.8)}
  \\
  
  \midrule

  Many-shot (1K) 
  & \textbf{61.9} & \textbf{88.1}{\tiny ($\pm$1.7)} & \underline{24.0}{\tiny ($\pm$1.5)} & \textbf{90.2}{\tiny ($\pm$0.0)} & \textbf{56.3}{\tiny ($\pm$0.3)} & \textbf{87.5}{\tiny ($\pm$0.2)} & \textbf{25.6}{\tiny ($\pm$0.6)} & \textbf{61.6}{\tiny ($\pm$1.2)}
  \\
  
  SITE 
  & 59.1 & 78.9{\tiny ($\pm$0.8)} & 23.5{\tiny ($\pm$0.7)} & 89.7{\tiny ($\pm$0.2)} & 55.5{\tiny ($\pm$0.1)} & 87.0{\tiny ($\pm$0.2)} & 21.8{\tiny ($\pm$0.1)} & 57.1{\tiny ($\pm$0.8)}
  \\

  ICV 
  & 59.0 & 80.1{\tiny ($\pm$0.6)} & \textbf{24.1}{\tiny ($\pm$1.3)} & 89.2{\tiny ($\pm$0.1)} & 55.8{\tiny ($\pm$0.1)} & 86.9{\tiny ($\pm$0.2)} & 21.4{\tiny ($\pm$0.2)} & 55.1{\tiny ($\pm$0.3)}
  \\
  
  LTV 
  & 58.2 & 78.4{\tiny ($\pm$3.2)} & 23.4{\tiny ($\pm$1.1)} & 88.9{\tiny ($\pm$0.6)} & 55.7{\tiny ($\pm$0.1)} & 86.9{\tiny ($\pm$0.2)} & 20.4{\tiny ($\pm$0.7)} & 53.9{\tiny ($\pm$1.4)}
  \\
  
  \textbf{AP} 
  & \underline{61.0} & \underline{84.0}{\tiny ($\pm$1.0)} & \textbf{24.1}{\tiny ($\pm$0.9)} & \underline{89.8}{\tiny ($\pm$0.1)} & \underline{56.0}{\tiny ($\pm$0.2)} & \underline{87.3}{\tiny ($\pm$0.2)} & \underline{24.9}{\tiny ($\pm$0.7)} & \underline{60.8}{\tiny ($\pm$0.9)}
  \\
  
  \midrule

  Many-shot (2K) 
  & \textbf{62.0} & \textbf{87.7}{\tiny ($\pm$0.9)} & 23.7{\tiny ($\pm$1.5)} & \textbf{90.2}{\tiny ($\pm$0.0)} & \textbf{56.3}{\tiny ($\pm$0.3)} & \textbf{87.7}{\tiny ($\pm$0.1)} & \textbf{27.1}{\tiny ($\pm$0.7)} & \textbf{61.4}{\tiny ($\pm$1.1)}
  \\
  
  SITE 
  & 59.1 & 78.8{\tiny ($\pm$0.9)} & 23.5{\tiny ($\pm$0.9)} & \underline{89.8}{\tiny ($\pm$0.1)} & 55.3{\tiny ($\pm$0.2)} & 86.9{\tiny ($\pm$0.3)} & 22.3{\tiny ($\pm$0.3)} & 56.9{\tiny ($\pm$0.4)} 
  \\

  ICV 
  & 58.9 & 80.1{\tiny ($\pm$0.4)} & \underline{24.2}{\tiny ($\pm$1.3)} & 89.2{\tiny ($\pm$0.2)} & 55.8{\tiny ($\pm$0.1)} & 86.9{\tiny ($\pm$0.2)} & 21.3{\tiny ($\pm$0.2)} & 55.0{\tiny ($\pm$0.3)}
  \\
  
  LTV 
  & 58.3 & 79.9{\tiny ($\pm$2.9)} & 23.5{\tiny ($\pm$1.3)} & 88.9{\tiny ($\pm$0.5)} & 55.8{\tiny ($\pm$0.1)} & 86.9{\tiny ($\pm$0.2)} & 19.3{\tiny ($\pm$0.9)} & 53.8{\tiny ($\pm$1.1)}
  \\
  
  \textbf{AP} 
  & \underline{61.3} & \underline{84.4}{\tiny ($\pm$3.7)} & \textbf{24.3}{\tiny ($\pm$1.1)} & 89.7{\tiny ($\pm$0.3)} & \underline{56.2}{\tiny ($\pm$0.3)} & \underline{87.5}{\tiny ($\pm$0.2)} & \underline{26.5}{\tiny ($\pm$0.6)} & \underline{60.6}{\tiny ($\pm$0.7)}
  \\
  
  \midrule

  Many-shot (4K) 
  & \textbf{62.2} & \textbf{88.1}{\tiny ($\pm$2.2)} & 24.0{\tiny ($\pm$0.8)} & \textbf{90.2}{\tiny ($\pm$0.0)} & \textbf{56.3}{\tiny ($\pm$0.3)} & \textbf{87.8}{\tiny ($\pm$0.1)} & \textbf{27.8}{\tiny ($\pm$0.7)} & \textbf{61.5}{\tiny ($\pm$1.3)}
  \\
  
  SITE 
  & 59.2 & 79.4{\tiny ($\pm$1.1)} & 23.1{\tiny ($\pm$0.7)} & \underline{89.8}{\tiny ($\pm$0.1)} & 55.6{\tiny ($\pm$0.4)} & 87.0{\tiny ($\pm$0.2)} & 22.9{\tiny ($\pm$0.2)} & 56.8{\tiny ($\pm$0.6)} 
  \\

  ICV 
  & 59.0 & 80.3{\tiny ($\pm$0.3)} & \textbf{24.3}{\tiny ($\pm$1.2)} & 89.2{\tiny ($\pm$0.1)} & \underline{55.7}{\tiny ($\pm$0.1)} & 86.9{\tiny ($\pm$0.2)} & 21.5{\tiny ($\pm$0.5)} & 55.3{\tiny ($\pm$0.5)}
  \\
  
  LTV 
  & 58.4 & 80.9{\tiny ($\pm$0.6)} & 23.8{\tiny ($\pm$1.2)} & 89.0{\tiny ($\pm$0.4)} & \underline{55.7}{\tiny ($\pm$0.2)} & 86.9{\tiny ($\pm$0.1)} & 18.5{\tiny ($\pm$1.3)} & 53.8{\tiny ($\pm$0.4)}
  \\
  
  \textbf{AP} 
  & \underline{61.0} & \underline{82.2}{\tiny ($\pm$1.9)} & \underline{24.1}{\tiny ($\pm$1.0)} & 89.6{\tiny ($\pm$0.2)} & \textbf{56.3}{\tiny ($\pm$0.2)} & \underline{87.5}{\tiny ($\pm$0.2)} & \underline{27.3}{\tiny ($\pm$0.6)} & \underline{59.6}{\tiny ($\pm$1.1)}
  \\
  
  \bottomrule
\end{tabular}}
\end{table}

\subsection{Multi-token Generation Task}

To examine whether the learned patches remain effective beyond single-token classification, we evaluate \ours\ on four autoregressive generation tasks using 32 calibration samples. GSM8K is evaluated using flexible-match accuracy, while DailyMail and CommonGen use ROUGE-L F1 \citep{lin-2004-rouge}. For FLORES, COMET \citep{rei-etal-2020-comet}, a neural metric for translation quality, is reported alongside ROUGE-L F1 to complement lexical overlap. We compare \ours\ against SITE, ICV, and LTV, whose original formulations support autoregressive generation. Other implicit ICL baselines primarily target single-token prediction and lack a standardized intervention procedure for multi-token decoding.

As shown in Table~\ref{tab:generation}, \ours\ generally outperforms the evaluated implicit ICL baselines across generation tasks and demonstration-context budgets. Averaged across the five context budgets on Llama, \ours\ achieves 70.3 accuracy on GSM8K, compared with 69.4 for SITE and 70.3 for explicit many-shot ICL. On FLORES Korean and Spanish, \ours\ obtains average ROUGE-L F1 scores of 19.9 and 52.0 and COMET scores of 76.5 and 84.6, respectively, outperforming SITE on both metrics. On DailyMail and CommonGen, \ours\ achieves average ROUGE-L F1 scores of 22.8 and 55.9 on Llama, and 26.2 and 60.3 on Gemma, respectively, approaching explicit many-shot ICL performance. These results suggest that the learned patches remain effective during autoregressive multi-token generation.

\section{Analysis}
\subsection{Inference Efficiency}
\begin{figure}[!b]
\centering
\includegraphics[width=1\linewidth]{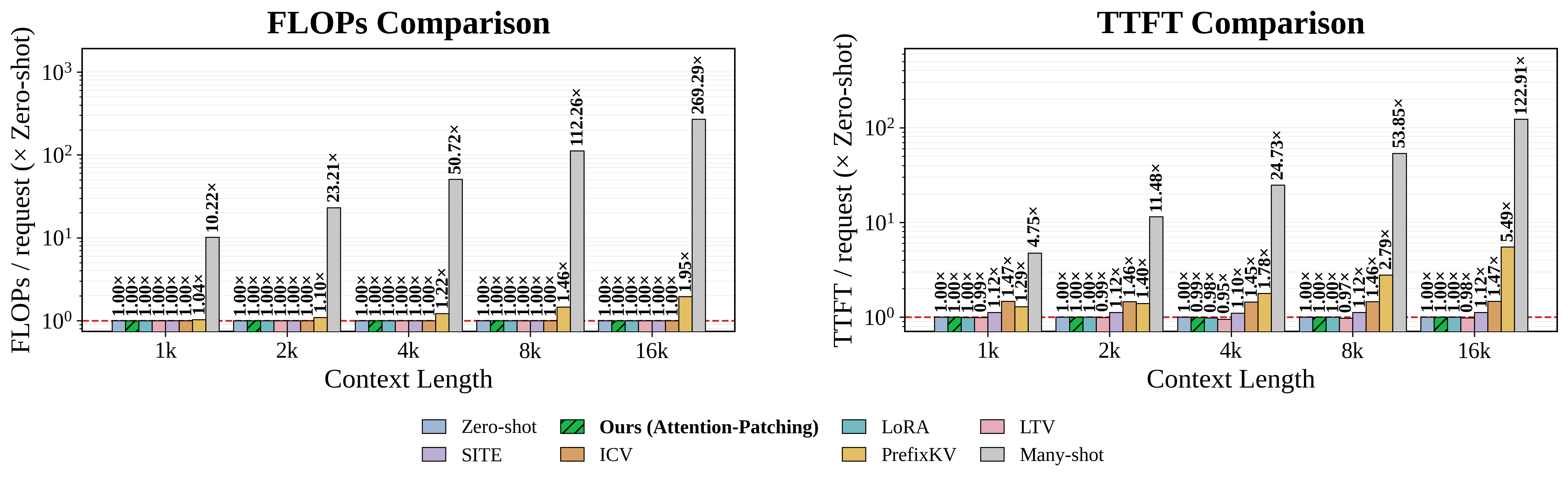}
\caption{Average inference FLOPs and TTFT on GSM8K using
Llama-3.2-3B-Instruct across different context lengths, averaged over 500 samples.}
\label{fig:efficiency}
\end{figure}

To investigate whether \ours\ preserves the inference efficiency, we measure its computational cost as the demonstration context used for calibration increases in length.

\paragraph{FLOPs.}
Figure~\ref{fig:efficiency} compares the per-query inference cost of \ours\ with zero-shot, LoRA, LTV, SITE, ICV, PrefixKV Caching, and explicit many-shot ICL on GSM8K across context lengths from 1K to 16K tokens. We estimate FLOPs using an analytical FLOP counter based on runtime tensor shapes. All results are normalized by the corresponding zero-shot result. As context length increases, the computational cost of explicit ICL rises from 10.22$\times$ to 269.29$\times$ that of zero-shot inference. In contrast, \ours\ maintains approximately the same per-query FLOPs as zero-shot inference, as the learned updates are incorporated directly into the existing attention output projections.


\paragraph{TTFT.}
A similar trend holds for time-to-first-token (TTFT). Many-shot ICL increases from $4.75\times$ to $122.91\times$, while \ours\ remains within $0.99$--$1.00\times$ of zero-shot inference. ICV also avoids processing demonstration tokens and has approximately constant FLOPs, but incurs a TTFT overhead of $1.45$--$1.47\times$, since each layer and token position requires adding a steering vector and renormalizing the activation. Together, the FLOPs and TTFT measurements show that \ours\ introduces negligible per-query inference overhead.

\subsection{Role of Linear and Bias Components}



To understand whether the gains of \ours\ stem from input-dependent transformations or fixed additive corrections, we compare linear, bias, and full affine patch formulations.
We evaluate two variants: Bias-only, parameterized by \(\Delta b^{(\ell)}\), and Weight-only, parameterized by \(Z^{(\ell)}\Delta W^{(\ell)}\).

\begin{table*}[!h]
\caption{
Component ablation of \ours\ on Llama-3.2-3B-Instruct. We report average accuracy across the 12 classification benchmarks.
}
\label{tab:patch_component}
\centering
\small
\setlength{\tabcolsep}{2pt}
\begin{tabular}{@{}l@{\hspace{1pt}}|@{\hspace{1pt}}cccccccccccc@{\hspace{1pt}}|@{\hspace{1pt}}c@{}}
  \toprule

  Method
  & AG & SST-2 & TREC & CSQA & PIQA & SST-5
  & MR & MRPC & CB & COPA & CREAK & AI2SciE
  & Avg. \\
  
  \midrule
  
  Bias
  & 58.8 & 87.1 & 68.7 & 76.1 & 75.8 & 34.2 
  & 89.0 & 68.2 & 58.2 & 93.8 & 77.2 & 87.0 
  & 72.8 \\
  
  Weight
  & 85.7 & 93.1 & 84.6 & 75.7 & 76.0 & 46.6 
  & 90.8 & 68.3 & 80.0 & 92.6 & 86.0 & 87.6 
  & 80.6 \\
  
  \textbf{Full}
  & 85.8 & 93.0 & 83.9 & 76.2 & 75.8 & 46.7 
  & 91.1 & 68.7 & 79.6 & 93.2 & 86.0 & 87.7 
  & 80.7 \\
  \bottomrule
\end{tabular}
\end{table*}


As shown in Table~\ref{tab:patch_component}, the weight component accounts for most of the performance of the full patch. On the classification benchmarks, Bias-only achieves an average accuracy of 72.8, whereas Weight-only reaches 80.6, nearly matching the 80.7 accuracy of the full Bias + Weight parameterization. 


From the adaptive steering perspective, $\Delta W^{(\ell)}$ combines multiple steering directions with input-dependent coefficients, whereas $\Delta b^{(\ell)}$ provides only a single fixed steering vector. This explains the smaller contribution of the bias component and suggests that the effectiveness of \ours\ primarily arises from input-dependent transformations rather than fixed additive corrections. 

\subsection{Singular Structure of Adaptive Steering}
\label{sec:singular_steering}

\begin{figure}[!h]
\centering
    \includegraphics[width=0.8\linewidth]{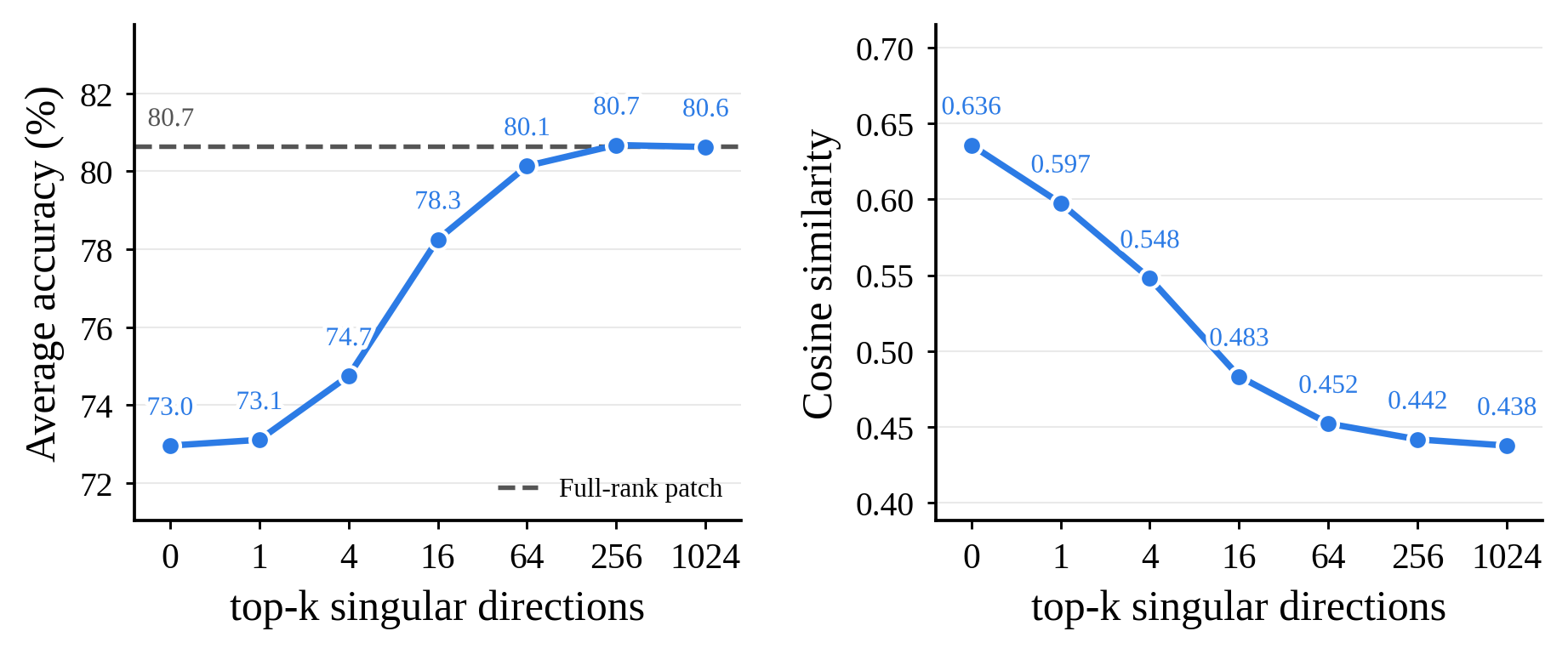}
\caption{Singular structure of adaptive steering on Llama-3.2-3B-Instruct. Average accuracy across 12 classification tasks (left) and cosine similarity to the bias-only steering vector averaged across tasks and layers (right) using rank-\(k\) truncated SVD of the weight patch.}
\label{fig:adaptive_steering_topk}
\end{figure}


To empirically examine the adaptive steering interpretation in Section~\ref{sec:adap_steering}, we compare the learned weight patch with a bias-only variant. For the bias-only variant, the closed-form solution is
\(
\Delta b_{\mathrm{bias}}^\top
=
\mu_{\Delta O}
=
\frac{1}{N}\sum_{j=1}^{N}\delta O_j,
\)
which corresponds to a fixed steering vector and matches the averaging form used in Contrastive Activation Addition \citep{rimsky-etal-2024-steering}.

We then replace $\Delta W_\star$ with its rank-$k$ truncated SVD,
\(
\Delta W_{\star,k}
=
\sum_{i=1}^{k}\sigma_i u_i v_i^\top,
\)
while retaining the bias $\Delta b_\star$ learned jointly with the full affine patch. The resulting correction is
\[
\widehat{\delta O_k^t}
=
\alpha
\left(
\sum_{i=1}^{k}
\sigma_i
\left(
Z_{\mathrm{non\text{-}ctx}}^t u_i
\right)
v_i^\top
+
\Delta b_\star^\top
\right).
\]
Unlike a fixed steering vector, the contribution of each singular direction is modulated by the current attention representation. Note that $k=0$ retains only the bias of the full affine patch and is therefore distinct from the separately fitted bias-only variant.


As shown in Figure~\ref{fig:adaptive_steering_topk}, the \(k=0\) intervention shows positive alignment with the bias-only steering vector, with a cosine similarity of 0.636. As the leading singular components are incorporated, the similarity steadily decreases to 0.438, while average accuracy increases from 73.0 to 80.7. This trend supports the adaptive steering view, where singular directions provide input-dependent refinements beyond a fixed steering vector.

\subsection{Layer-wise Structure of Learned Patches}
\label{sec:layer_wise_structure}

\begin{figure}[t]
\centering
\includegraphics[width=1\linewidth]{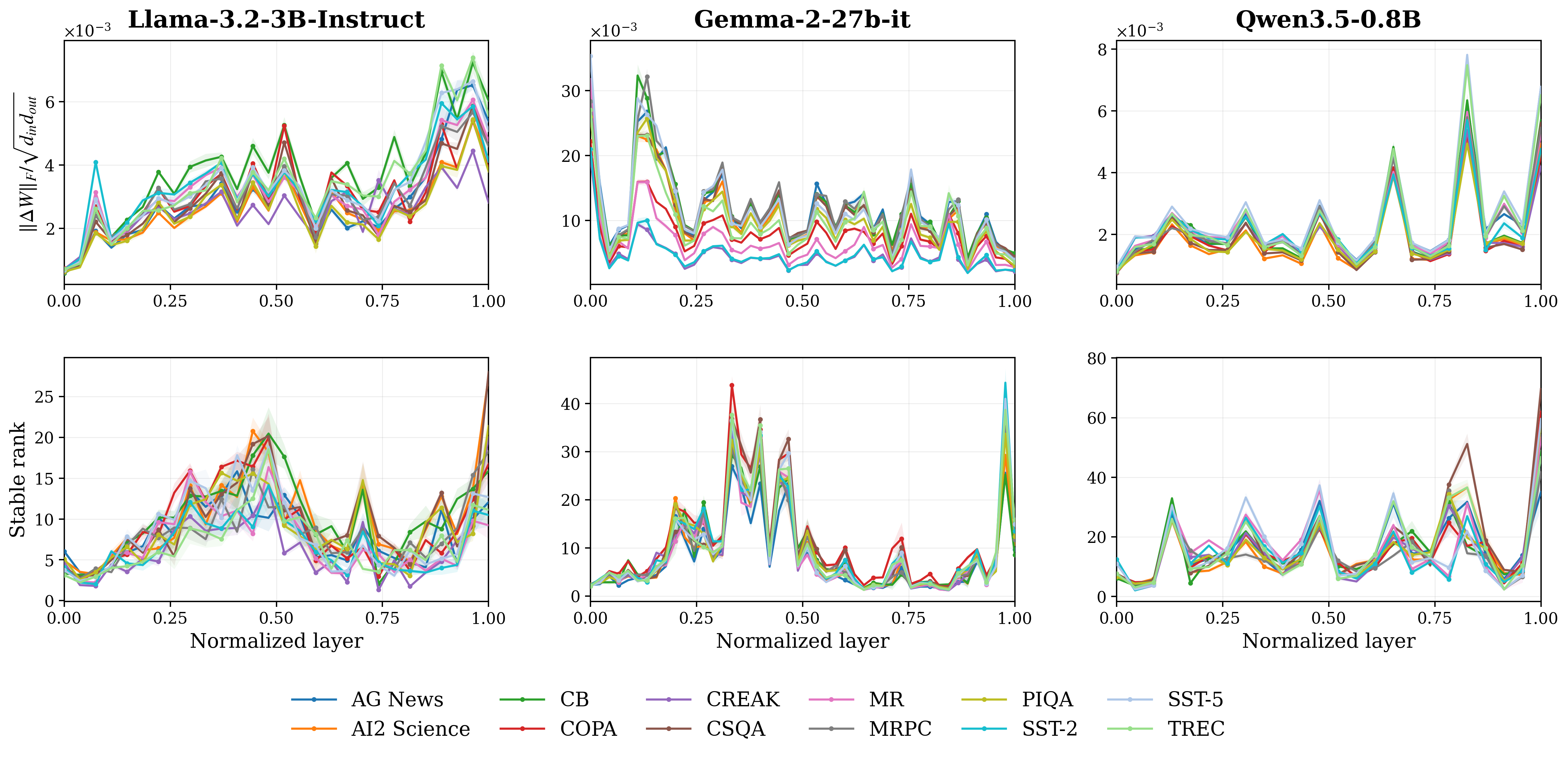}
\caption{Layer-wise structure of the learned attention patches across classification benchmarks. The top row shows the normalized Frobenius norm of the patch weights, while the bottom row shows their stable rank. Each column corresponds to a model architecture, and each curve represents one classification task.}
\label{fig:layer_patching}
\end{figure}



%

Figure~\ref{fig:layer_patching} illustrates the layer-wise structure of the learned patches across different model architectures, characterizing their update magnitude and effective dimensionality using the normalized Frobenius norm and stable rank, respectively.

For each layer \(\ell\), the top row of Figure~\ref{fig:layer_patching} reports the update magnitude using the normalized Frobenius norm: \( m^{(\ell)} = \frac{\lVert \Delta W^{(\ell)} \rVert_F} {\sqrt{d_{\mathrm{in}}d_{\mathrm{out}}}} \), and
the bottom row reports the stable rank of each matrix update:
\(
\operatorname{srank}\!\left(\Delta W^{(\ell)}\right)
=
\frac{\left\|\Delta W^{(\ell)}\right\|_F^2}
{\left\|\Delta W^{(\ell)}\right\|_2^2}
=
\frac{\sum_i \left(\sigma_i^{(\ell)}\right)^2}
{\left(\max_i \sigma_i^{(\ell)}\right)^2},
\)
where \(\sigma_i^{(\ell)}\) denotes the \(i\)-th singular value of \(\Delta W^{(\ell)}\).

The two measures exhibit distinct layer-wise patterns across model architectures, indicating that larger updates do not necessarily correspond to higher-dimensional transformations. This suggests that \ours\ distributes task-specific corrections across different layers and singular directions rather than relying on a uniform intervention throughout the model.

\subsection{Performance Alignment Between ICL and \ours}

\begin{figure}
    \centering
    \includegraphics[width=1\linewidth]{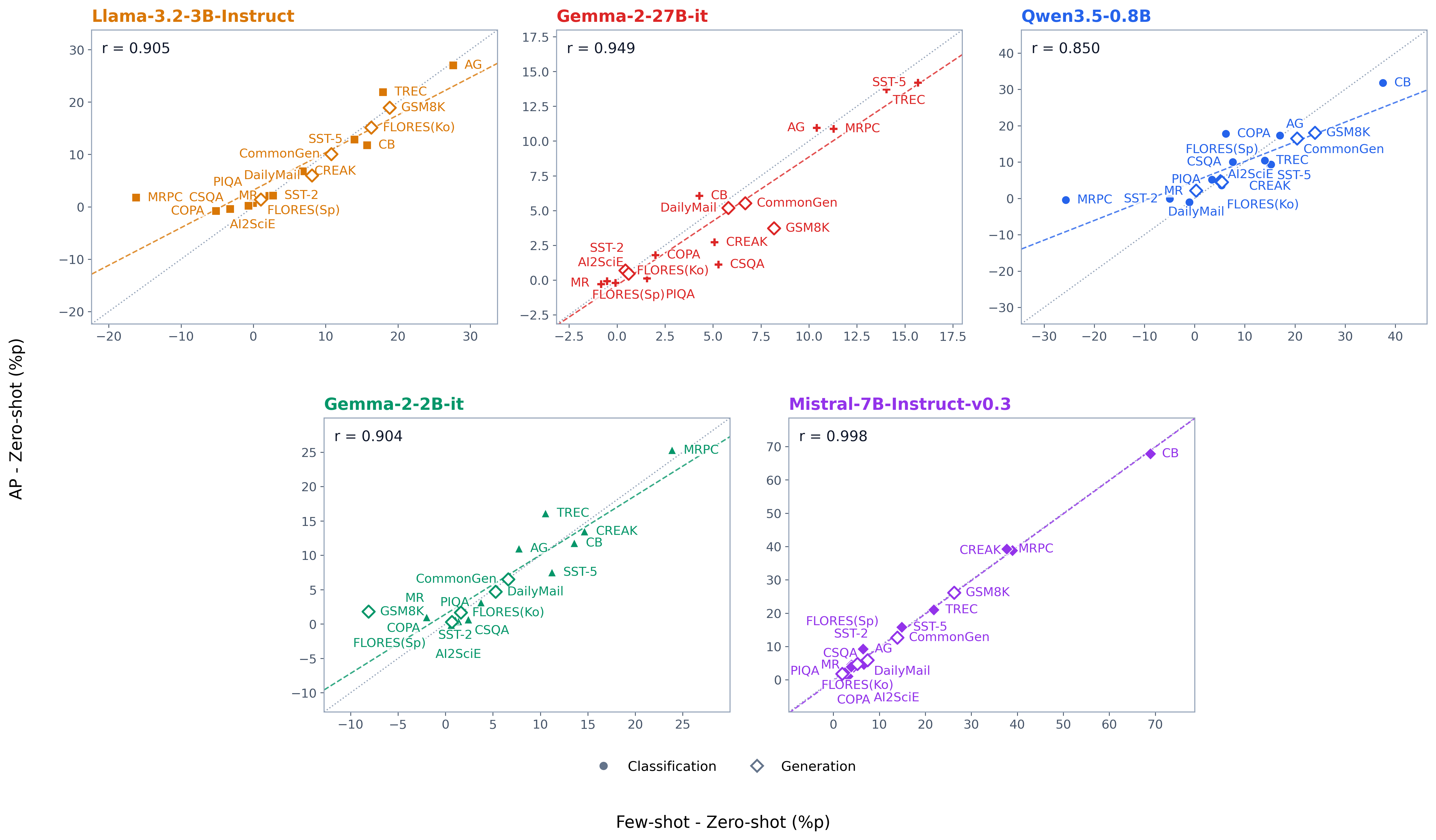}
    \caption{
    Correlation between performance changes induced by
    explicit ICL and AP relative to zero-shot inference.
    Each point represents a task.}
    \label{fig:icl_ap_correlation}
\end{figure}





To investigate whether AP captures the task-specific behavior of explicit ICL, we analyze the relationship between the performance changes induced by explicit ICL and AP. For each task and model, we compute the performance differences relative to zero-shot inference: $\mathrm{ICL} - \text{Zero-shot}$ and $\mathrm{AP} - \text{Zero-shot}$. As shown in Figure~\ref{fig:icl_ap_correlation}, the two performance differences exhibit strong positive correlations across all five models, with Pearson correlation coefficients ranging from 0.850 to 0.998. These results suggest that AP closely follows the performance trends of explicit ICL.

\section{Conclusion}
We introduce \ours, a method that compresses the effects of in-context demonstrations into attention patches, enabling query-only inference without repeatedly processing the demonstrations.
Across 12 classification benchmarks, \ours\ achieved the strongest average performance among the evaluated implicit ICL methods, while remaining competitive with explicit ICL. \ours\ also generally outperformed the evaluated implicit ICL methods on multi-token generation tasks across context budgets from 1K to 16K tokens, demonstrating its applicability beyond single-token classification.
These results suggest \ours\ as an effective approach for retaining demonstration-induced behavior while reducing the inference-time overhead of explicit ICL.
\section{Discussion and Limitations}

\ours\ enables backpropagation-free adaptation by capturing demonstration-induced effects from internal model representations and incorporating them directly into attention projections without additional inference-time modules. However, this approach requires access to internal activations and model weights, limiting its applicability to open-weight models rather than closed-source models accessible only through black-box APIs.

\subsection*{AI use statement}

In this work, we used generative AI tools to assist with translation, improve the readability and clarity of the manuscript, assist in the writing of proofs, and assist with code implementation. All AI-assisted content was reviewed and verified by the authors, who take full responsibility for the final content of this work.                                  We have not used generative AI tools for developing theoretical models or conceptual frameworks, formulating mathematical claims, providing critical ingredients for proving mathematical claims, proposing or refining hypotheses, designing or providing feedback on research methodology or experiments, cleaning and reformatting datasets, or interpreting results. Generating synthetic datasets and supporting qualitative and thematic data analysis were not applicable to this work.

\subsection*{Reproducibility statement}

We provide the source code and configurations for all key experiments, including model training and evaluation. Detailed descriptions of the proposed method, experimental settings, and implementation details are provided in Section~\ref{sec:experimental_setup} and Appendix~\ref{appendix:reproducibility} to facilitate reproducibility.


\bibliography{ref}
\bibliographystyle{iclr2027_conference}

\appendix
\section{Theoretical Analysis of \ours}
\subsection{Derivation of the Closed-Form Ridge Solution}
\label{app:closed_form}

In this section, we derive the closed-form solution used to compute the layer-wise attention patch in \ours. For conciseness, we omit the layer index \(\ell\) throughout the derivation.

We collect the non-contextual attention representations and their target demonstration effects over all query-answer token positions in the calibration set as
\[
\mathbf{Z}
\in
\mathbb{R}^{N\times d_{\mathrm{in}}},
\qquad
\mathbf{\Delta O}
\in
\mathbb{R}^{N\times d_{\mathrm{out}}},
\]
where \(N\) is the total number of query-answer token positions. The rows of \(\mathbf{Z}\) are taken from \(Z_{\mathrm{non\text{-}ctx}}\), and the corresponding rows of \(\mathbf{\Delta O}\) are the demonstration effects \(\delta O\) defined in Section~\ref{sec:demo_effect}.

The patch parameters
\[
\Delta W
\in
\mathbb{R}^{d_{\mathrm{in}}\times d_{\mathrm{out}}},
\qquad
\Delta b
\in
\mathbb{R}^{d_{\mathrm{out}}}
\]
are fitted by
\[
\min_{\Delta W,\Delta b}
\mathcal{L}(\Delta W,\Delta b)
=
\frac{1}{N}
\left\|
\mathbf{Z}\Delta W
+
\mathbf{1}_{N}\Delta b^{\top}
-
\mathbf{\Delta O}
\right\|_{F}^{2}
+
\rho\left\|\Delta W\right\|_{F}^{2},
\]
where \(\mathbf{1}_{N}\) is the all-ones vector and \(\rho>0\) is the ridge coefficient.

We first solve for the bias term. Differentiating the objective with respect to \(\Delta b\) gives
\[
\nabla_{\Delta b}\mathcal{L}
=
\frac{2}{N}
\left(
\mathbf{Z}\Delta W
+
\mathbf{1}_{N}\Delta b^{\top}
-
\mathbf{\Delta O}
\right)^{\top}
\mathbf{1}_{N}.
\]
Setting the gradient to zero yields
\[
\mathbf{1}_{N}^{\top}\mathbf{Z}\Delta W
+
N\Delta b^{\top}
-
\mathbf{1}_{N}^{\top}\mathbf{\Delta O}
=
0.
\]

We define the row means as
\[
\mu_Z
=
\frac{1}{N}\mathbf{1}_{N}^{\top}\mathbf{Z},
\qquad
\mu_{\Delta O}
=
\frac{1}{N}\mathbf{1}_{N}^{\top}\mathbf{\Delta O}.
\]
For any fixed \(\Delta W\), the optimal bias is therefore
\[
\Delta b^{\top}
=
\mu_{\Delta O}
-
\mu_Z\Delta W.
\]

We then center the two matrices as
\[
\overline{\mathbf{Z}}
=
\mathbf{Z}
-
\mathbf{1}_{N}\mu_Z,
\qquad
\overline{\mathbf{\Delta O}}
=
\mathbf{\Delta O}
-
\mathbf{1}_{N}\mu_{\Delta O}.
\]
Substituting the expression for \(\Delta b\) into the residual gives
\[
\begin{aligned}
\mathbf{Z}\Delta W
+
\mathbf{1}_{N}\Delta b^{\top}
-
\mathbf{\Delta O}
&=
\mathbf{Z}\Delta W
+
\mathbf{1}_{N}
\left(
\mu_{\Delta O}
-
\mu_Z\Delta W
\right)
-
\mathbf{\Delta O}
\\
&=
\left(
\mathbf{Z}
-
\mathbf{1}_{N}\mu_Z
\right)\Delta W
-
\left(
\mathbf{\Delta O}
-
\mathbf{1}_{N}\mu_{\Delta O}
\right)
\\
&=
\overline{\mathbf{Z}}\Delta W
-
\overline{\mathbf{\Delta O}}.
\end{aligned}
\]
The objective therefore reduces to
\[
\mathcal{L}(\Delta W)
=
\frac{1}{N}
\left\|
\overline{\mathbf{Z}}\Delta W
-
\overline{\mathbf{\Delta O}}
\right\|_{F}^{2}
+
\rho\left\|\Delta W\right\|_{F}^{2}.
\]

Differentiating with respect to \(\Delta W\) gives
\[
\nabla_{\Delta W}\mathcal{L}
=
\frac{2}{N}
\overline{\mathbf{Z}}^{\top}
\left(
\overline{\mathbf{Z}}\Delta W
-
\overline{\mathbf{\Delta O}}
\right)
+
2\rho\Delta W.
\]
Setting the gradient to zero and rearranging gives
\[
\left(
\overline{\mathbf{Z}}^{\top}\overline{\mathbf{Z}}
+
N\rho I_{d_{\mathrm{in}}}
\right)\Delta W
=
\overline{\mathbf{Z}}^{\top}
\overline{\mathbf{\Delta O}}.
\]
Since \(\rho>0\), the matrix on the left is positive definite and therefore invertible. The optimal weight patch is
\[
\Delta W_\star
=
\left(
\overline{\mathbf{Z}}^{\top}\overline{\mathbf{Z}}
+
N\rho I_{d_{\mathrm{in}}}
\right)^{-1}
\overline{\mathbf{Z}}^{\top}
\overline{\mathbf{\Delta O}},
\]
with the corresponding bias
\[
\Delta b_\star^{\top}
=
\mu_{\Delta O}
-
\mu_Z\Delta W_\star.
\]

Restoring the layer index, the calibrated patch parameters are
\[
\Delta W_\star^{(\ell)}
=
\left(
\overline{\mathbf{Z}}^{(\ell)\top}
\overline{\mathbf{Z}}^{(\ell)}
+
N\rho I_{d_{\mathrm{in}}}
\right)^{-1}
\overline{\mathbf{Z}}^{(\ell)\top}
\overline{\mathbf{\Delta O}}^{(\ell)},
\]
and
\[
\Delta b_\star^{(\ell)\top}
=
\mu_{\Delta O}^{(\ell)}
-
\mu_Z^{(\ell)}
\Delta W_\star^{(\ell)}.
\]

\subsection{Functional View}
\label{app:func_view}

Consider a single attention head and a target position \(t\) in a query-answer pair, and omit the layer and head indices for clarity. Let \(\mathcal X_t\) denote the source positions from the query-answer pair visible to \(t\), and let \(\mathcal D_t\) denote the demonstration-token positions visible to \(t\) in the contextual prompt. The non-contextual attention representation is
\[
z_t^{\mathrm{non\text{-}ctx}}
=
\frac{
\sum_{s\in\mathcal X_t}
\exp(a_{ts}^{\mathrm{non\text{-}ctx}})
v_s^{\mathrm{non\text{-}ctx}}
}{
\sum_{s\in\mathcal X_t}
\exp(a_{ts}^{\mathrm{non\text{-}ctx}})
}.
\]

When demonstrations are added, the attention weights and value representations of the original context may change. We define the contextual attention mass over \(\mathcal X_t\) and the corresponding normalized representation as
\[
A_t^{\mathrm{ctx}}
=
\sum_{s\in\mathcal X_t}
\exp(a_{ts}^{\mathrm{ctx}}),
\qquad
z_{t,\mathcal X}^{\mathrm{ctx}}
=
\frac{1}{A_t^{\mathrm{ctx}}}
\sum_{s\in\mathcal X_t}
\exp(a_{ts}^{\mathrm{ctx}})
v_s^{\mathrm{ctx}}.
\]
For the demonstration-token positions, we similarly define
\[
B_t
=
\sum_{m\in\mathcal D_t}
\exp(a_{tm}^{\mathrm{ctx}}),
\qquad
r_{D,t}
=
\frac{1}{B_t}
\sum_{m\in\mathcal D_t}
\exp(a_{tm}^{\mathrm{ctx}})
v_m^{\mathrm{ctx}}.
\]

The contextual attention representation can then be written as
\[
z_t^{\mathrm{ctx}}
=
(1-\lambda_t)z_{t,\mathcal X}^{\mathrm{ctx}}
+
\lambda_t r_{D,t},
\qquad
\lambda_t
=
\frac{B_t}{A_t^{\mathrm{ctx}}+B_t}.
\]
Subtracting the non-contextual output gives
\[
z_t^{\mathrm{ctx}}
-
z_t^{\mathrm{non\text{-}ctx}}
=
\underbrace{
\lambda_t r_{D,t}
}_{\text{demonstration retrieval}}
-
\underbrace{
\lambda_t z_t^{\mathrm{non\text{-}ctx}}
}_{\text{attention redistribution}}
+
\underbrace{
(1-\lambda_t)
\left(
z_{t,\mathcal X}^{\mathrm{ctx}}
-
z_t^{\mathrm{non\text{-}ctx}}
\right)
}_{\text{context change}}.
\]

This decomposition shows that the effect of demonstrations is not a single fixed shift. At each token position, the head-level changes are combined across attention heads to form \(Z_{\mathrm{ctx}}^{t}\) and \(Z_{\mathrm{non\text{-}ctx}}^{t}\). Over all token positions, these representations give the layer-wise demonstration effect
\[
\delta O
=
\left(
Z_{\mathrm{ctx}}
-
Z_{\mathrm{non\text{-}ctx}}
\right)
W_O.
\]
From this perspective, \ours\ can be viewed as compactly representing these demonstration-induced changes through an affine correction, which provides the basis for the adaptive steering view.

\subsection{pseudocode}
\label{app:pseudocode}

Algorithm~\ref{alg:patching} describes the calibration and patch strength selection procedure of \ours. We first collect contextual attention outputs from the model. Patches are then fitted sequentially using ridge regression. Each fitted patch remains active when fitting subsequent layers, so later layers receive
activations that reflect the effects of previously fitted patches. Finally, we select the patch strength on a held-out validation set.

\begin{algorithm*}[t]
\caption{\ours}
\label{alg:patching}

\textbf{Input}: Model $f_{\theta}$,
demonstrations $\mathcal D$,
calibration set $\mathcal Q=\{(x_i,y_i)\}_{i=1}^{M}$,
validation set $\mathcal V$ \\
\textbf{Parameter}: Ridge coefficient $\rho$,
patch strength candidates $\mathcal A$ \\
\textbf{Output}: Patched model $f_{\tilde{\theta}}$

\begin{algorithmic}[1]

\STATE Initialize $P\leftarrow\emptyset$

\STATE \textit{// Collect contextual attention outputs}
\FOR{each $(x_i,y_i)\in\mathcal Q$}
    \STATE $O_\mathrm{ctx}^{(1:L)}(x_i,y_i)
    \leftarrow \operatorname{AttnOut}
    (f_\theta([\mathcal D;x_i;y_i]))$
\ENDFOR

\STATE \textit{// Sequential fitting}
\FOR{$\ell=1,\ldots,L$}
    \FOR{each $(x_i,y_i)\in\mathcal Q$}
        \STATE $(Z_\mathrm{non\text{-}ctx}^{(\ell)}(x_i,y_i),
        O_\mathrm{non\text{-}ctx}^{(\ell)}(x_i,y_i))
        \leftarrow \operatorname{AttnProj}_{\ell}
        (f_{\theta+P}([x_i;y_i]))$
        \STATE $\delta O^{(\ell)}(x_i,y_i)
        \leftarrow O_\mathrm{ctx}^{(\ell)}(x_i,y_i)
        -O_\mathrm{non\text{-}ctx}^{(\ell)}(x_i,y_i)$
    \ENDFOR

    \STATE $\mathbf{Z}^{(\ell)}
    \leftarrow \operatorname{RowConcat}_i
    Z_{\mathrm{non\text{-}ctx}}^{(\ell)}(x_i,y_i)$
    \STATE $\boldsymbol{\Delta O}^{(\ell)}
    \leftarrow \operatorname{RowConcat}_i
    \delta O^{(\ell)}(x_i,y_i)$

    \STATE \textit{// Fit the attention patch via ridge regression}
    \STATE \makebox[\linewidth][c]{$
    (\Delta W_*^{(\ell)},\Delta b_*^{(\ell)})
    \leftarrow \arg\min_{\Delta W,\Delta b}
    \frac{1}{N}
    \left\|
    \mathbf{Z}^{(\ell)}\Delta W+
    \mathbf{1}_N\Delta b^\top-
    \boldsymbol{\Delta O}^{(\ell)}
    \right\|_F^2
    +\rho\|\Delta W\|_F^2
    $}

    \STATE $P\leftarrow P\cup
    \{(\ell,\Delta W_*^{(\ell)},\Delta b_*^{(\ell)})\}$
\ENDFOR

\STATE \textit{// Patch strength selection}
\STATE $\alpha_*\leftarrow
\arg\max_{\alpha\in\mathcal A}
\operatorname{Score}(f_{\theta+\alpha P},\mathcal V)$

\STATE $\tilde{\theta}\leftarrow\theta+\alpha_*P$

\STATE \textbf{return} $f_{\tilde{\theta}}$

\end{algorithmic}
\end{algorithm*}

\section{Experimental Setup}
\subsection{Reproducibility Details}


\label{appendix:reproducibility}

All classification experiments are repeated using five random seeds, $\{42,43,44,45,46\}$, and the results are averaged across the five runs. All experiments are conducted on NVIDIA RTX 6000 Ada Generation GPUs.

For each classification task, we use a fixed demonstration set containing five examples per class and construct a disjoint calibration set from the training split. \ours\ uses 25 calibration examples per class. We select the patch strength \(\alpha\) by sweeping \(\alpha \in \{0.001, 0.1, 0.2, \ldots, 1.0 \}\) over 100 held-out validation instances and choosing the value that achieves the best validation performance. For evaluation, we use up to 500 instances from the evaluation split of each classification benchmark, using all available instances for smaller datasets.

For multi-token generation, we use 32 calibration examples and follow the same procedure for selecting the patch strength \(\alpha\). For evaluation, we use 500 instances for each task, with demonstration-context budgets ranging from 1K to 16K tokens; for models with shorter context windows, we use budgets from 1K to 4K tokens.

\begin{figure}[!h]
\centering
\includegraphics[width=1\linewidth]{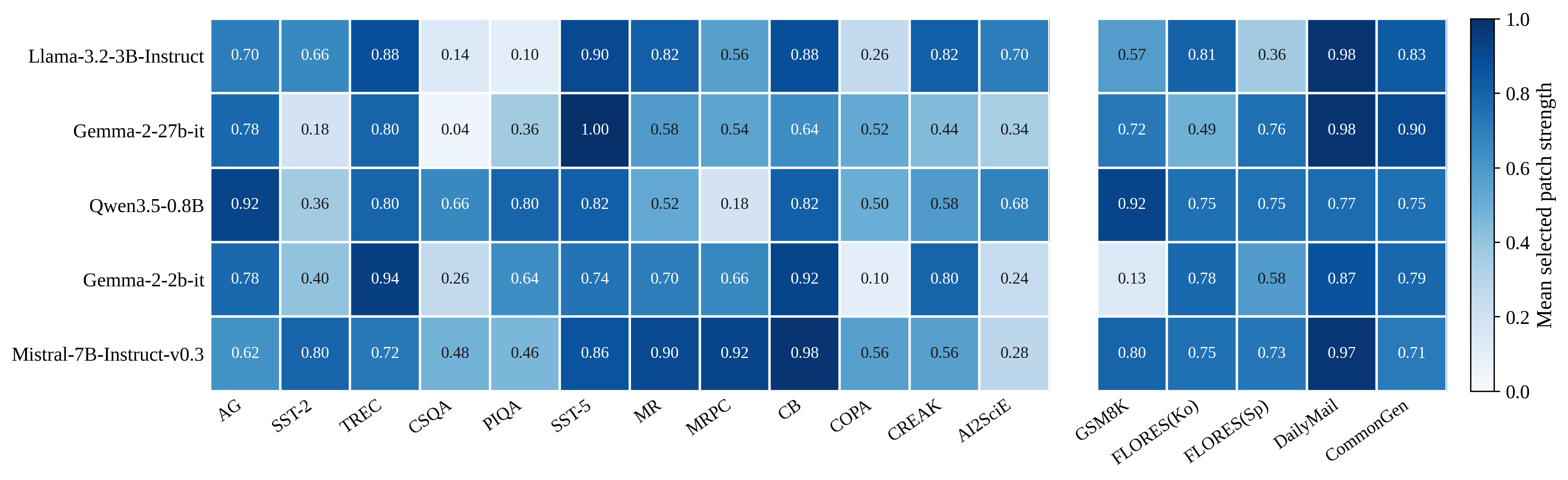}
\caption{Selected patch strength \(\alpha\) for each task and model. The left panel corresponds to classification tasks, where values are averaged across five independent runs, and the right panel corresponds to multi-token generation tasks, where values are averaged across the evaluated context lengths and three independent runs for each model. In each experiment, \(\alpha\) is selected on the held-out validation set following the procedure described above.
}
\label{fig:selected_patch_strength}
\end{figure}

Figure~\ref{fig:selected_patch_strength} reports the selected patch strengths across tasks and models. 

\begin{table*}[!t]
\caption{Prompt templates used in the classification experiments.}
\label{tab:prompt_templates}
\centering
\begin{tabular}{@{}p{0.12\linewidth}p{0.21\linewidth}p{0.60\linewidth}@{}}
\toprule
\textbf{Dataset} & \textbf{Task Type} & \textbf{Template} \\
\midrule

AGNews
& Topic classification
& News: \{text\}; \newline
Type: [World, Sports, Business, Technology] \\

\midrule

SST-2
& Sentiment
& Review: \{text\}; \newline
Sentiment: [negative, positive] \\

\midrule

TREC
& Question type classification
& Question: \{text\}; \newline
Answer Type: [Abbreviation, Entity, Description, Person, Location, Number] \\

\midrule

CSQA
& Commonsense
& Question: \{question\}; A. \{optA\}; B. \{optB\}; C. \{optC\}; D. \{optD\}; E. \{optE\}; \newline
Answer (A/B/C/D/E); Options: [A, B, C, D, E] \\

\midrule

PIQA
& Physical commonsense
& Goal: \{goal\}; A. \{optA\}; B. \{optB\}; \newline
Answer (A/B); Options: [A, B] \\

\midrule

SST-5
& Sentiment 
& Sentence: \{text\}; \newline
Sentiment: [terrible, negative, neutral, positive, great] \\

\midrule

MR
& Movie review
& Review: \{text\}; \newline
Sentiment: [negative, positive] \\

\midrule

MRPC
& Paraphrase
& \{pair\}; A. Paraphrase; B. Not paraphrase; \newline
Answer (A/B); Options: [A, B] \\

\midrule

CB
& NLI 
& \{pair\}; A. Entailment; B. Contradiction; C. Neutral; \newline
Answer (A/B/C); Options: [A, B, C] \\

\midrule

CREAK
& Claim verification
& Claim: \{claim\}; Label: yes / no; \newline
Options: [yes, no] \\

\midrule

COPA
& Causal reasoning 
& \{context\}; A. \{optA\}; B. \{optB\}; \newline
Answer (A/B); Options: [A, B] \\

\midrule

AI2SciE
& Science MCQ 
& Question: \{question\}; A. \{optA\}; B. \{optB\}; C. \{optC\}; D. \{optD\}; \newline
E. \{optE\}; F. \{optF\}; G. \{optG\}; H. \{optH\}; \newline
Answer (A/B/C/\ldots); Options: [A, B, C, D, E, F, G, H] \\

\bottomrule
\end{tabular}
\end{table*}
\subsection{Classification Prompt Templates}
Table~\ref{tab:prompt_templates} summarizes the dataset-specific prompt templates used in our classification experiments. Each template specifies the input format and the corresponding set of candidate labels.

\subsection{LoRA Setup}

For comparison with \ours, we apply LoRA to the attention output projection (\texttt{o\_proj}), corresponding to the projection modified by our method. We use a rank of 8, a scaling factor of 16, and a dropout rate of 0.05. The model is trained for one epoch using AdamW with a learning rate of $2\times10^{-4}$ and a batch size of 4. We use no weight decay or learning-rate warmup and set the maximum gradient norm to 1.0.

For classification, LoRA is trained on 5, 15 or 25 examples per class, corresponding to the data budgets reported in the experiments. For multi-token generation, we use 32 training examples, matching the calibration-set size used for Attention-Patching.


\section{Additional Results}
\subsection{Classification Results on Additional Models}
\label{appendix:one_token_prediction}

\begin{table*}[!t]
\caption{
Baseline comparison across classification benchmarks.
All results are averaged over five independent runs.
The best and second-best results in each column within each model
are indicated in bold and underline, respectively.
}
\label{appendix:tab:classification}
\centering
\resizebox{\textwidth}{!}{%
\begin{tabular}{@{}lc|*{12}{c}@{}}
 \toprule

Method & Avg. & AG & SST-2 & TREC & CSQA & PIQA & SST-5 & MR & MRPC & CB & COPA & CREAK & AI2SciE \\
\midrule
\multicolumn{14}{c}{\textbf{Gemma-2-2b-it}} \\
\midrule

0-shot 
& 72.7 & 75.6 & 92.6 & 72.8 & 69.8 & 71.8 & 36.8 & 91.6 & 45.3 & 62.5 & 94.0 & 72.4 & 87.4 
\\

ICL 
& \underline{80.1} & 83.3{\tiny ($\pm$2.4)} & \textbf{93.6}{\tiny ($\pm$1.2)} & \underline{83.3}{\tiny ($\pm$3.0)} & \textbf{72.2}{\tiny ($\pm$0.9)} & \textbf{75.5}{\tiny ($\pm$1.0)} & \textbf{48.0}{\tiny ($\pm$1.8)} & \textbf{92.2}{\tiny ($\pm$0.7)} & \underline{69.2}{\tiny ($\pm$4.1)} & \textbf{76.1}{\tiny ($\pm$9.8)} & 92.0{\tiny ($\pm$1.4)} & \textbf{87.0}{\tiny ($\pm$1.1)} & \underline{88.8}{\tiny ($\pm$0.6)} 
\\

\midrule

LoRA 
& 75.2 & \underline{87.0}{\tiny ($\pm$2.3)} & 91.6{\tiny ($\pm$1.5)} & 81.7{\tiny ($\pm$5.5)} & 68.0{\tiny ($\pm$1.3)} & \underline{75.2}{\tiny ($\pm$0.6)} & 44.6{\tiny ($\pm$1.7)} & 91.0{\tiny ($\pm$1.8)} & 45.2{\tiny ($\pm$11.4)} & 71.4{\tiny ($\pm$4.6)} & 94.4{\tiny ($\pm$0.5)} & 64.8{\tiny ($\pm$2.2)} & 87.9{\tiny ($\pm$0.2)} 
\\

SITE 
& 78.8 & \textbf{89.6}{\tiny ($\pm$1.2)} & 93.0{\tiny ($\pm$0.4)} & 82.9{\tiny ($\pm$2.5)} & 69.0{\tiny ($\pm$1.9)} & 74.5{\tiny ($\pm$1.5)} & 36.0{\tiny ($\pm$1.3)} & 91.3{\tiny ($\pm$0.7)} & 68.4{\tiny ($\pm$0.0)} & 73.2{\tiny ($\pm$5.6)} & 94.6{\tiny ($\pm$0.5)} & \underline{86.4}{\tiny ($\pm$2.3)} & 87.0{\tiny ($\pm$1.3)} 
\\

\midrule

TV 
& 73.7 & 85.4{\tiny ($\pm$1.0)} & 92.5{\tiny ($\pm$0.1)} & 76.4{\tiny ($\pm$0.1)} & 70.3{\tiny ($\pm$0.3)} & 71.8{\tiny ($\pm$0.3)} & 37.7{\tiny ($\pm$1.2)} & 90.0{\tiny ($\pm$1.8)} & 43.5{\tiny ($\pm$3.3)} & 57.1{\tiny ($\pm$4.0)} & 93.4{\tiny ($\pm$0.9)} & 79.3{\tiny ($\pm$4.0)} & 86.6{\tiny ($\pm$1.2)} 
\\

I2CL 
& 73.0 & 82.8{\tiny ($\pm$1.0)} & 86.8{\tiny ($\pm$7.3)} & 75.8{\tiny ($\pm$8.6)} & 66.2{\tiny ($\pm$2.7)} & 73.4{\tiny ($\pm$1.8)} & 36.4{\tiny ($\pm$2.9)} & 89.4{\tiny ($\pm$0.5)} & 64.0{\tiny ($\pm$6.8)} & 61.1{\tiny ($\pm$3.4)} & 93.0{\tiny ($\pm$1.2)} & 62.5{\tiny ($\pm$8.3)} & 85.1{\tiny ($\pm$1.3)} 
\\

FV 
& 70.6 & 75.2{\tiny ($\pm$1.4)} & 79.9{\tiny ($\pm$1.3)} & 78.7{\tiny ($\pm$2.0)} & 65.8{\tiny ($\pm$0.3)} & 69.8{\tiny ($\pm$1.5)} & 29.8{\tiny ($\pm$1.0)} & 75.5{\tiny ($\pm$0.6)} & \underline{69.2}{\tiny ($\pm$0.4)} & 72.1{\tiny ($\pm$0.5)} & 90.6{\tiny ($\pm$0.8)} & 55.6{\tiny ($\pm$0.6)} & 84.6{\tiny ($\pm$0.7)} 
\\

IV 
& 73.0 & 78.2{\tiny ($\pm$0.8)} & 90.4{\tiny ($\pm$0.9)} & 75.9{\tiny ($\pm$0.9)} & 66.5{\tiny ($\pm$0.6)} & 59.8{\tiny ($\pm$1.0)} & \underline{45.5}{\tiny ($\pm$1.8)} & 87.2{\tiny ($\pm$1.2)} & 65.7{\tiny ($\pm$2.5)} & 59.3{\tiny ($\pm$1.1)} & 85.8{\tiny ($\pm$0.9)} & 77.6{\tiny ($\pm$2.2)} & 84.6{\tiny ($\pm$1.2)} 
\\

ICV 
& 72.7 & 75.6{\tiny ($\pm$0.0)} & 92.1{\tiny ($\pm$1.1)} & 72.8{\tiny ($\pm$0.0)} & 69.6{\tiny ($\pm$0.7)} & 72.9{\tiny ($\pm$1.1)} & 38.5{\tiny ($\pm$0.5)} & \underline{91.8}{\tiny ($\pm$0.4)} & 47.3{\tiny ($\pm$2.5)} & 59.3{\tiny ($\pm$0.8)} & \underline{94.8}{\tiny ($\pm$0.8)} & 70.7{\tiny ($\pm$1.2)} & 86.7{\tiny ($\pm$0.3)} 
\\

LTV 
& 77.1 & 81.8{\tiny ($\pm$3.0)} & 93.1{\tiny ($\pm$1.1)} & 78.6{\tiny ($\pm$5.2)} & 69.0{\tiny ($\pm$1.5)} & 71.1{\tiny ($\pm$1.2)} & 43.2{\tiny ($\pm$3.5)} & 90.2{\tiny ($\pm$1.3)} & 62.7{\tiny ($\pm$4.3)} & 71.8{\tiny ($\pm$6.8)} & 91.8{\tiny ($\pm$1.3)} & 83.0{\tiny ($\pm$1.1)} & \textbf{89.0}{\tiny ($\pm$0.8)} 
\\

\midrule

TP 
& 66.5 & 51.8{\tiny ($\pm$2.6)} & 89.0{\tiny ($\pm$0.9)} & 47.5{\tiny ($\pm$7.2)} & 54.8{\tiny ($\pm$1.5)} & 69.6{\tiny ($\pm$1.2)} & 34.8{\tiny ($\pm$1.3)} & 87.7{\tiny ($\pm$1.3)} & 68.3{\tiny ($\pm$0.4)} & 55.7{\tiny ($\pm$8.5)} & 92.0{\tiny ($\pm$1.4)} & 81.0{\tiny ($\pm$2.6)} & 65.8{\tiny ($\pm$5.4)} 
\\

\textbf{AP (Ours)} 
& \textbf{80.3} & 86.6{\tiny ($\pm$2.1)} & \underline{93.2}{\tiny ($\pm$0.3)} & \textbf{88.9}{\tiny ($\pm$1.1)} & \underline{70.5}{\tiny ($\pm$0.2)} & 74.9{\tiny ($\pm$1.2)} & 44.3{\tiny ($\pm$4.7)} & 91.5{\tiny ($\pm$0.7)} & \textbf{70.6}{\tiny ($\pm$0.9)} & \underline{74.3}{\tiny ($\pm$7.2)} & \textbf{95.0}{\tiny ($\pm$0.7)} & 85.9{\tiny ($\pm$0.8)} & 87.8{\tiny ($\pm$0.6)} 
\\

\midrule
\multicolumn{14}{c}{\textbf{Mistral-7B-Instruct-v0.3}} \\
\midrule

0-shot 
& 65.0 & 76.6 & 90.4 & 58.8 & 69.6 & 77.0 & 36.2 & 89.2 & 32.6 & 19.6 & 92.0 & 51.6 & 86.8 
\\

ICL 
& \textbf{82.8} & 83.5{\tiny ($\pm$1.8)} & \underline{94.0}{\tiny ($\pm$1.0)} & 80.9{\tiny ($\pm$4.0)} & \textbf{76.2}{\tiny ($\pm$0.7)} & \textbf{80.0}{\tiny ($\pm$0.6)} & \underline{51.1}{\tiny ($\pm$1.5)} & \underline{93.1}{\tiny ($\pm$0.4)} & 69.6{\tiny ($\pm$5.2)} & 88.2{\tiny ($\pm$5.5)} & \textbf{96.6}{\tiny ($\pm$1.1)} & \textbf{90.7}{\tiny ($\pm$0.4)} & \textbf{90.0}{\tiny ($\pm$0.4)} 
\\

\midrule

LoRA 
& 77.3 & \underline{86.9}{\tiny ($\pm$2.9)} & 90.4{\tiny ($\pm$0.7)} & \textbf{86.7}{\tiny ($\pm$3.4)} & 69.8{\tiny ($\pm$0.8)} & 78.2{\tiny ($\pm$1.1)} & 46.2{\tiny ($\pm$3.7)} & 89.0{\tiny ($\pm$2.8)} & 51.5{\tiny ($\pm$19.9)} & 71.4{\tiny ($\pm$15.3)} & 92.8{\tiny ($\pm$0.8)} & 77.2{\tiny ($\pm$2.5)} & 87.3{\tiny ($\pm$0.6)} 
\\

SITE 
& 79.9 & \textbf{89.4}{\tiny ($\pm$2.2)} & 93.8{\tiny ($\pm$0.6)} & 75.5{\tiny ($\pm$4.9)} & 70.1{\tiny ($\pm$1.7)} & 75.8{\tiny ($\pm$3.4)} & 42.2{\tiny ($\pm$4.2)} & 91.8{\tiny ($\pm$0.6)} & 68.4{\tiny ($\pm$0.0)} & 81.8{\tiny ($\pm$3.9)} & 93.4{\tiny ($\pm$0.5)} & 88.6{\tiny ($\pm$1.2)} & 87.8{\tiny ($\pm$0.7)} 
\\

\midrule

TV 
& 67.7 & 81.6{\tiny ($\pm$4.2)} & 88.2{\tiny ($\pm$3.4)} & 61.7{\tiny ($\pm$5.4)} & 71.4{\tiny ($\pm$0.4)} & 77.4{\tiny ($\pm$1.5)} & 36.5{\tiny ($\pm$0.9)} & 92.1{\tiny ($\pm$0.6)} & 35.8{\tiny ($\pm$9.2)} & 30.4{\tiny ($\pm$3.3)} & 93.2{\tiny ($\pm$1.1)} & 57.0{\tiny ($\pm$3.9)} & 87.1{\tiny ($\pm$0.8)} 
\\

I2CL 
& 77.2 & 85.1{\tiny ($\pm$1.6)} & 90.4{\tiny ($\pm$1.3)} & 76.7{\tiny ($\pm$8.8)} & 64.8{\tiny ($\pm$2.6)} & 77.0{\tiny ($\pm$1.0)} & 44.6{\tiny ($\pm$2.1)} & 88.2{\tiny ($\pm$1.8)} & 63.3{\tiny ($\pm$10.6)} & 73.2{\tiny ($\pm$3.3)} & 93.6{\tiny ($\pm$1.1)} & 81.4{\tiny ($\pm$2.5)} & 87.6{\tiny ($\pm$1.5)} 
\\

FV 
& 81.4 & 85.7{\tiny ($\pm$0.8)} & 93.9{\tiny ($\pm$0.5)} & 62.3{\tiny ($\pm$0.7)} & \underline{74.4}{\tiny ($\pm$1.2)} & \underline{79.9}{\tiny ($\pm$0.6)} & 43.4{\tiny ($\pm$1.3)} & 92.4{\tiny ($\pm$0.7)} & \underline{73.7}{\tiny ($\pm$0.7)} & \textbf{95.5}{\tiny ($\pm$0.5)} & 96.1{\tiny ($\pm$0.4)} & 90.0{\tiny ($\pm$0.8)} & \underline{89.4}{\tiny ($\pm$0.7)} 
\\

IV 
& 81.9 & 81.4{\tiny ($\pm$0.4)} & 92.1{\tiny ($\pm$0.8)} & \underline{84.4}{\tiny ($\pm$1.4)} & 73.4{\tiny ($\pm$0.9)} & 79.5{\tiny ($\pm$0.8)} & 40.3{\tiny ($\pm$1.2)} & 91.0{\tiny ($\pm$0.7)} & \textbf{73.9}{\tiny ($\pm$1.2)} & \underline{93.5}{\tiny ($\pm$1.1)} & 95.2{\tiny ($\pm$0.6)} & 89.8{\tiny ($\pm$0.5)} & 88.6{\tiny ($\pm$0.6)} 
\\

ICV 
& 69.0 & 76.6{\tiny ($\pm$0.0)} & 90.1{\tiny ($\pm$0.3)} & 64.4{\tiny ($\pm$0.8)} & 71.0{\tiny ($\pm$0.1)} & 76.8{\tiny ($\pm$0.0)} & 36.0{\tiny ($\pm$0.2)} & 90.4{\tiny ($\pm$0.7)} & 32.6{\tiny ($\pm$0.0)} & 27.1{\tiny ($\pm$8.8)} & 92.6{\tiny ($\pm$0.5)} & 83.3{\tiny ($\pm$1.5)} & 86.9{\tiny ($\pm$0.2)} 
\\

LTV 
& 79.6 & 81.7{\tiny ($\pm$1.4)} & 92.0{\tiny ($\pm$1.2)} & 79.8{\tiny ($\pm$3.2)} & 69.2{\tiny ($\pm$1.3)} & 77.0{\tiny ($\pm$1.5)} & 50.0{\tiny ($\pm$1.0)} & 89.5{\tiny ($\pm$1.2)} & 68.5{\tiny ($\pm$4.5)} & 82.5{\tiny ($\pm$2.0)} & 93.2{\tiny ($\pm$1.3)} & 84.4{\tiny ($\pm$2.9)} & 88.0{\tiny ($\pm$0.9)} 
\\

\midrule

TP 
& 77.2 & 69.5{\tiny ($\pm$5.3)} & 90.6{\tiny ($\pm$0.6)} & 68.9{\tiny ($\pm$1.9)} & 71.0{\tiny ($\pm$0.9)} & 78.8{\tiny ($\pm$1.1)} & 39.4{\tiny ($\pm$1.6)} & 91.1{\tiny ($\pm$1.0)} & 68.1{\tiny ($\pm$1.5)} & 78.2{\tiny ($\pm$3.4)} & \underline{96.2}{\tiny ($\pm$0.8)} & 89.0{\tiny ($\pm$0.8)} & 85.3{\tiny ($\pm$0.4)} 
\\

\textbf{AP (Ours)} 
& \underline{82.7} & 85.8{\tiny ($\pm$1.8)} & \textbf{94.3}{\tiny ($\pm$0.5)} & 79.8{\tiny ($\pm$3.4)} & 74.2{\tiny ($\pm$0.4)} & 79.6{\tiny ($\pm$0.7)} & \textbf{52.0}{\tiny ($\pm$2.2)} & \textbf{93.7}{\tiny ($\pm$0.6)} & 71.9{\tiny ($\pm$2.7)} & 87.5{\tiny ($\pm$3.1)} & 95.8{\tiny ($\pm$2.3)} & \underline{90.4}{\tiny ($\pm$1.0)} & 87.9{\tiny ($\pm$0.6)} 
\\
\bottomrule
\end{tabular}}
\end{table*}

Table \ref{appendix:tab:classification} presents additional classification results on Gemma-2-2b-it and Mistral-7B-Instruct-v0.3, complementing the main results reported in Section \ref{main_result}.

\subsection{Multi-token Generation Results on Additional Models}

\begin{table}[!t]
\caption{Multi-token generation results under different context lengths, averaged over three independent runs. We report flexible-match accuracy on GSM8K, ROUGE-L F1 and COMET scores on FLORES, and ROUGE-L F1 scores on DailyMail and CommonGen. The best and second-best results in each column within each model and context length are indicated in bold and underline, respectively.}
\renewcommand{\arraystretch}{0.72}
\scriptsize
\label{append:tab:generation}
\centering

\resizebox{\columnwidth}{!}{%
\begin{tabular}{l c| c cc cc cc}
  \toprule
  \multirow{2}{*}{Method}
  & \multirow{2}{*}{Avg.}
  & \multirow{2}{*}{GSM8K}
  & \multicolumn{2}{c}{FLORES(Ko)}
  & \multicolumn{2}{c}{FLORES(Sp)}
  & \multirow{2}{*}{DailyMail}
  & \multirow{2}{*}{CommonGen} \\
  \cmidrule(lr){4-5}
  \cmidrule(lr){6-7}
  & & & ROUGE-L F1 & COMET & ROUGE-L F1 & COMET & & \\
  \midrule

  \multicolumn{9}{c}{\textbf{Gemma-2-2b-it}} \\
  \midrule

  Zero-shot 
  & 51.7 & 55.2{\tiny ($\pm$1.2)} & 20.9{\tiny ($\pm$0.7)} & 81.5{\tiny ($\pm$0.1)} & 52.7{\tiny ($\pm$0.4)} & 85.5{\tiny ($\pm$0.3)} & 18.7{\tiny ($\pm$0.2)} & 47.3{\tiny ($\pm$0.8)} 
  \\
  
  LoRA 
  & 51.9 & 57.2{\tiny ($\pm$1.7)} & 21.1{\tiny ($\pm$0.6)} & 80.9{\tiny ($\pm$0.4)} & 52.8{\tiny ($\pm$0.3)} & 85.8{\tiny ($\pm$0.2)} & 15.9{\tiny ($\pm$0.3)} & 49.4{\tiny ($\pm$1.2)}
  \\

  \midrule
  
  Many-shot (1K) 
  & \underline{53.2} & 48.5{\tiny ($\pm$3.1)} & \underline{22.6}{\tiny ($\pm$0.6)} & \textbf{84.4}{\tiny ($\pm$0.4)} & \textbf{53.4}{\tiny ($\pm$0.2)} & \textbf{85.9}{\tiny ($\pm$0.3)} & \underline{23.4}{\tiny ($\pm$0.6)} & \underline{54.4}{\tiny ($\pm$1.3)}
  \\
  
  SITE 
  & 50.8 & 45.9{\tiny ($\pm$1.5)} & 21.6{\tiny ($\pm$0.8)} & 83.3{\tiny ($\pm$0.4)} & 51.5{\tiny ($\pm$0.2)} & 85.4{\tiny ($\pm$0.2)} & 18.4{\tiny ($\pm$1.0)} & 49.8{\tiny ($\pm$0.9)}
  \\
  
  ICV 
  & 52.0 & \underline{56.5}{\tiny ($\pm$0.9)} & 21.3{\tiny ($\pm$0.3)} & 81.2{\tiny ($\pm$0.6)} & 52.8{\tiny ($\pm$0.3)} & 85.5{\tiny ($\pm$0.3)} & 18.9{\tiny ($\pm$0.5)} & 47.6{\tiny ($\pm$0.7)}
  \\
  
  LTV 
  & 42.7 & 26.8{\tiny ($\pm$14.0)} & 20.3{\tiny ($\pm$0.5)} & 79.6{\tiny ($\pm$0.6)} & 51.2{\tiny ($\pm$0.6)} & 84.6{\tiny ($\pm$0.1)} & 19.8{\tiny ($\pm$0.7)} & 16.3{\tiny ($\pm$4.9)}
  \\
  
  \textbf{AP} 
  & \textbf{54.4} & \textbf{57.6}{\tiny ($\pm$0.2)} & \textbf{23.2}{\tiny ($\pm$1.0)} & \underline{83.4}{\tiny ($\pm$0.9)} & \underline{53.2}{\tiny ($\pm$0.1)} & \underline{85.7}{\tiny ($\pm$0.3)} & \textbf{23.5}{\tiny ($\pm$0.8)} & \textbf{54.5}{\tiny ($\pm$1.1)}
  \\

  \midrule
  
  Many-shot (2K) 
  & \underline{53.1} & 47.5{\tiny ($\pm$2.0)} & \underline{22.2}{\tiny ($\pm$0.7)} & \textbf{84.5}{\tiny ($\pm$0.4)} & \textbf{53.4}{\tiny ($\pm$0.1)} & \textbf{86.0}{\tiny ($\pm$0.2)} & \textbf{24.0}{\tiny ($\pm$0.3)} & \underline{54.1}{\tiny ($\pm$1.4)}
  \\
  
  SITE 
  & 50.9 & 46.7{\tiny ($\pm$2.4)} & 21.5{\tiny ($\pm$0.5)} & 83.4{\tiny ($\pm$0.1)} & 51.7{\tiny ($\pm$0.5)} & 85.4{\tiny ($\pm$0.3)} & 17.8{\tiny ($\pm$0.5)} & 49.6{\tiny ($\pm$0.4)}
  \\
  
  ICV 
  & 52.0 & \underline{56.5}{\tiny ($\pm$1.0)} & 21.2{\tiny ($\pm$0.5)} & 81.6{\tiny ($\pm$0.1)} & \underline{52.8}{\tiny ($\pm$0.3)} & 85.5{\tiny ($\pm$0.3)} & 18.9{\tiny ($\pm$0.5)} & 47.6{\tiny ($\pm$0.7)}
  \\
  
  LTV 
  & 42.1 & 23.5{\tiny ($\pm$1.5)} & 20.5{\tiny ($\pm$0.6)} & 79.7{\tiny ($\pm$0.5)} & 50.9{\tiny ($\pm$0.7)} & 84.5{\tiny ($\pm$0.1)} & 19.8{\tiny ($\pm$1.1)} & 15.5{\tiny ($\pm$3.4)}
  \\
  
  \textbf{AP} 
  & \textbf{54.2} & \textbf{57.4}{\tiny ($\pm$2.1)} & \textbf{22.5}{\tiny ($\pm$1.3)} & \underline{83.5}{\tiny ($\pm$0.4)} & \underline{52.8}{\tiny ($\pm$0.5)} & \underline{85.6}{\tiny ($\pm$0.3)} & \underline{23.3}{\tiny ($\pm$0.2)} & \textbf{54.2}{\tiny ($\pm$1.6)}
  \\

  \midrule
  
  Many-shot (4K) 
  & \underline{52.8} & 45.3{\tiny ($\pm$1.7)} & \textbf{22.7}{\tiny ($\pm$1.1)} & \textbf{84.3}{\tiny ($\pm$0.3)} & \textbf{53.4}{\tiny ($\pm$0.3)} & \textbf{86.0}{\tiny ($\pm$0.2)} & \textbf{24.6}{\tiny ($\pm$0.2)} & \textbf{53.1}{\tiny ($\pm$1.5)}
  \\
  
  SITE 
  & 50.7 & 45.9{\tiny ($\pm$0.9)} & 21.8{\tiny ($\pm$1.2)} & \underline{83.2}{\tiny ($\pm$0.4)} & 51.5{\tiny ($\pm$0.3)} & 85.4{\tiny ($\pm$0.4)} & 17.6{\tiny ($\pm$0.3)} & 49.7{\tiny ($\pm$0.6)}
  \\
  
  ICV 
  & 51.9 & \textbf{56.5}{\tiny ($\pm$1.3)} & 21.0{\tiny ($\pm$0.4)} & 81.2{\tiny ($\pm$0.5)} & 52.8{\tiny ($\pm$0.3)} & 85.5{\tiny ($\pm$0.3)} & 19.0{\tiny ($\pm$0.4)} & 47.6{\tiny ($\pm$0.7)}
  \\
  
  LTV 
  & 42.0 & 23.8{\tiny ($\pm$5.9)} & 20.6{\tiny ($\pm$0.9)} & 80.0{\tiny ($\pm$1.2)} & 51.1{\tiny ($\pm$0.6)} & 84.7{\tiny ($\pm$0.1)} & 18.3{\tiny ($\pm$1.4)} & 15.8{\tiny ($\pm$3.9)}
  \\
  
  \textbf{AP} 
  & \textbf{53.7} & \underline{56.0}{\tiny ($\pm$0.7)} & \underline{22.1}{\tiny ($\pm$1.3)} & 83.0{\tiny ($\pm$0.3)} & \underline{53.0}{\tiny ($\pm$0.2)} & \underline{85.7}{\tiny ($\pm$0.3)} & \underline{23.5}{\tiny ($\pm$0.1)} & \underline{52.6}{\tiny ($\pm$0.7)}
  \\
  
  \midrule
  
  \multicolumn{9}{c}{\textbf{Mistral-7B-Instruct-v0.3}} \\
  \midrule

  Zero-shot 
  & 45.7 & 25.6{\tiny ($\pm$0.6)} & 18.1{\tiny ($\pm$0.8)} & 75.2{\tiny ($\pm$0.7)} & 51.0{\tiny ($\pm$0.5)} & 84.0{\tiny ($\pm$0.1)} & 20.4{\tiny ($\pm$0.2)} & 45.3{\tiny ($\pm$0.7)}
  \\
  
  LoRA 
  & 49.8 & 35.4{\tiny ($\pm$2.8)} & 21.4{\tiny ($\pm$0.9)} & 80.1{\tiny ($\pm$0.5)} & 51.0{\tiny ($\pm$0.2)} & 85.0{\tiny ($\pm$0.1)} & 25.6{\tiny ($\pm$0.9)} & 50.0{\tiny ($\pm$1.1)}
  \\

  \midrule

  Many-shot (1K) 
  & \textbf{54.5} & \textbf{54.1}{\tiny ($\pm$0.6)} & \underline{22.8}{\tiny ($\pm$0.6)} & \textbf{81.5}{\tiny ($\pm$0.6)} & \textbf{53.0}{\tiny ($\pm$0.3)} & \textbf{85.7}{\tiny ($\pm$0.2)} & \textbf{25.8}{\tiny ($\pm$1.2)} & \textbf{58.4}{\tiny ($\pm$0.7)}
  \\
  
  SITE 
  & 50.4 & 41.7{\tiny ($\pm$4.0)} & 20.5{\tiny ($\pm$0.4)} & 78.2{\tiny ($\pm$0.2)} & \underline{52.1}{\tiny ($\pm$0.3)} & 84.8{\tiny ($\pm$0.4)} & 20.9{\tiny ($\pm$0.4)} & 54.8{\tiny ($\pm$0.9)}
  \\
  
  ICV 
  & 46.0 & 25.8{\tiny ($\pm$0.5)} & 19.1{\tiny ($\pm$0.4)} & 75.5{\tiny ($\pm$0.8)} & 51.0{\tiny ($\pm$0.5)} & 84.0{\tiny ($\pm$0.1)} & 21.0{\tiny ($\pm$0.3)} & 45.4{\tiny ($\pm$0.9)}
  \\
  
  LTV 
  & 37.9 & 21.0{\tiny ($\pm$2.3)} & 10.7{\tiny ($\pm$3.1)} & 54.4{\tiny ($\pm$2.8)} & 49.5{\tiny ($\pm$0.4)} & 82.6{\tiny ($\pm$0.5)} & 17.5{\tiny ($\pm$4.7)} & 29.9{\tiny ($\pm$4.6)}
  \\
  
  \textbf{AP} 
  & \underline{54.0} & \underline{53.5}{\tiny ($\pm$2.6)} & \textbf{23.1}{\tiny ($\pm$0.8)} & \underline{80.4}{\tiny ($\pm$0.1)} & \textbf{53.0}{\tiny ($\pm$0.1)} & \underline{85.4}{\tiny ($\pm$0.2)} & \underline{24.8}{\tiny ($\pm$0.8)} & \underline{57.8}{\tiny ($\pm$0.8)}
  \\

  \midrule
  
  Many-shot (2K) 
  & \textbf{54.8} & \textbf{52.8}{\tiny ($\pm$0.7)} & \textbf{23.0}{\tiny ($\pm$1.6)} & \textbf{81.8}{\tiny ($\pm$0.3)} & \textbf{53.0}{\tiny ($\pm$0.1)} & \textbf{85.9}{\tiny ($\pm$0.2)} & \textbf{28.2}{\tiny ($\pm$0.8)} & \textbf{59.1}{\tiny ($\pm$1.2)}
  \\
  
  SITE 
  & 50.2 & 40.1{\tiny ($\pm$5.9)} & 21.0{\tiny ($\pm$0.5)} & 78.4{\tiny ($\pm$0.4)} & \underline{52.1}{\tiny ($\pm$0.3)} & 84.8{\tiny ($\pm$0.3)} & 20.6{\tiny ($\pm$0.4)} & 54.8{\tiny ($\pm$0.6)}
  \\
  
  ICV 
  & 46.0 & 26.1{\tiny ($\pm$0.5)} & 19.1{\tiny ($\pm$0.2)} & 75.8{\tiny ($\pm$0.5)} & 50.9{\tiny ($\pm$0.5)} & 84.0{\tiny ($\pm$0.2)} & 21.0{\tiny ($\pm$0.3)} & 45.3{\tiny ($\pm$0.8)}
  \\
  
  LTV 
  & 38.1 & 21.3{\tiny ($\pm$2.4)} & 10.3{\tiny ($\pm$2.9)} & 54.1{\tiny ($\pm$1.7)} & 49.8{\tiny ($\pm$0.7)} & 82.7{\tiny ($\pm$0.7)} & 18.9{\tiny ($\pm$2.8)} & 29.8{\tiny ($\pm$1.4)} 
  \\
  
  \textbf{AP} 
  & \underline{54.0} & \underline{51.9}{\tiny ($\pm$0.6)} & \underline{22.5}{\tiny ($\pm$1.2)} & \underline{81.0}{\tiny ($\pm$0.4)} & \textbf{53.0}{\tiny ($\pm$0.4)} & \underline{85.6}{\tiny ($\pm$0.2)} & \underline{26.7}{\tiny ($\pm$0.9)} & \underline{57.4}{\tiny ($\pm$0.6)}
  \\

  \midrule
  
  Many-shot (4K) 
  & \textbf{54.7} & \underline{50.9}{\tiny ($\pm$1.0)} & \textbf{23.4}{\tiny ($\pm$0.9)} & \textbf{81.5}{\tiny ($\pm$0.5)} & \textbf{52.9}{\tiny ($\pm$0.1)} & \textbf{85.9}{\tiny ($\pm$0.1)} & \textbf{28.5}{\tiny ($\pm$0.5)} & \textbf{59.8}{\tiny ($\pm$1.6)}
  \\
  
  SITE 
  & 50.2 & 39.9{\tiny ($\pm$6.2)} & 21.1{\tiny ($\pm$0.9)} & 78.3{\tiny ($\pm$0.2)} & 51.9{\tiny ($\pm$0.2)} & 84.8{\tiny ($\pm$0.3)} & 20.5{\tiny ($\pm$0.3)} & 54.6{\tiny ($\pm$0.6)}
  \\

  ICV 
  & 46.2 & 27.1{\tiny ($\pm$0.3)} & 19.1{\tiny ($\pm$0.5)} & 75.7{\tiny ($\pm$0.5)} & 51.0{\tiny ($\pm$0.5)} & 84.0{\tiny ($\pm$0.2)} & 21.0{\tiny ($\pm$0.4)} & 45.2{\tiny ($\pm$0.7)}
  \\
  
  LTV 
  & 39.0 & 22.3{\tiny ($\pm$2.8)} & 10.9{\tiny ($\pm$2.5)} & 57.2{\tiny ($\pm$2.5)} & 49.5{\tiny ($\pm$0.6)} & 82.5{\tiny ($\pm$0.7)} & 20.5{\tiny ($\pm$0.6)} & 30.1{\tiny ($\pm$3.5)}
  \\
  
  \textbf{AP} 
  & \underline{54.3} & \textbf{52.9}{\tiny ($\pm$1.9)} & \underline{22.7}{\tiny ($\pm$0.5)} & \underline{80.5}{\tiny ($\pm$1.2)} & \underline{52.7}{\tiny ($\pm$0.3)} & \underline{85.6}{\tiny ($\pm$0.1)} & \underline{26.7}{\tiny ($\pm$0.6)} & \underline{58.7}{\tiny ($\pm$0.6)}
  \\

  \midrule
  
  Many-shot (8K) 
  & \textbf{54.9} & \textbf{52.3}{\tiny ($\pm$1.1)} & \textbf{23.7}{\tiny ($\pm$0.9)} & \textbf{81.9}{\tiny ($\pm$0.8)} & \textbf{53.0}{\tiny ($\pm$0.3)} & \textbf{86.0}{\tiny ($\pm$0.1)} & \textbf{28.2}{\tiny ($\pm$0.4)} & \textbf{58.9}{\tiny ($\pm$1.0)}
  \\
  
  SITE 
  & 50.4 & 40.8{\tiny ($\pm$2.1)} & 21.3{\tiny ($\pm$0.2)} & 78.4{\tiny ($\pm$0.3)} & 52.0{\tiny ($\pm$0.2)} & 84.9{\tiny ($\pm$0.3)} & 20.6{\tiny ($\pm$0.3)} & 54.9{\tiny ($\pm$1.0)}
  \\
  
  ICV 
  & 46.1 & 26.3{\tiny ($\pm$0.2)} & 19.2{\tiny ($\pm$0.4)} & 75.6{\tiny ($\pm$0.8)} & 51.1{\tiny ($\pm$0.5)} & 84.0{\tiny ($\pm$0.1)} & 21.0{\tiny ($\pm$0.5)} & 45.4{\tiny ($\pm$0.8)}
  \\
  
  LTV 
  & 39.2 & 21.5{\tiny ($\pm$3.4)} & 11.6{\tiny ($\pm$2.5)} & 58.4{\tiny ($\pm$2.1)} & 49.6{\tiny ($\pm$0.7)} & 82.7{\tiny ($\pm$0.9)} & 20.9{\tiny ($\pm$0.2)} & 29.6{\tiny ($\pm$0.7)} 
  \\
  
  \textbf{AP} 
  & \underline{54.0} & \underline{51.7}{\tiny ($\pm$1.2)} & \underline{22.9}{\tiny ($\pm$1.9)} & \underline{80.4}{\tiny ($\pm$0.7)} & \underline{52.7}{\tiny ($\pm$0.1)} & \underline{85.6}{\tiny ($\pm$0.1)} & \underline{26.7}{\tiny ($\pm$1.0)} & \underline{57.8}{\tiny ($\pm$0.6)}
  \\

  \midrule
  
  Many-shot (16K) 
  & \textbf{54.6} & \textbf{49.2}{\tiny ($\pm$0.9)} & \textbf{24.1}{\tiny ($\pm$0.6)} & \textbf{81.9}{\tiny ($\pm$0.5)} & \textbf{52.9}{\tiny ($\pm$0.1)} & \textbf{86.0}{\tiny ($\pm$0.1)} & \textbf{28.5}{\tiny ($\pm$0.5)} & \textbf{59.9}{\tiny ($\pm$1.3)}
  \\
  
  SITE 
  & 50.7 & 43.1{\tiny ($\pm$5.9)} & 21.1{\tiny ($\pm$0.3)} & 78.5{\tiny ($\pm$0.1)} & 52.1{\tiny ($\pm$0.2)} & 85.0{\tiny ($\pm$0.3)} & 20.4{\tiny ($\pm$0.3)} & 54.5{\tiny ($\pm$0.4)}
  \\
  
  ICV 
  & 46.1 & 26.5{\tiny ($\pm$0.5)} & 19.2{\tiny ($\pm$0.5)} & 75.6{\tiny ($\pm$0.9)} & 51.1{\tiny ($\pm$0.4)} & 84.1{\tiny ($\pm$0.1)} & 20.9{\tiny ($\pm$0.3)} & 45.4{\tiny ($\pm$0.7)} 
  \\
  
  LTV 
  & 39.0 & 21.3{\tiny ($\pm$1.5)} & 12.2{\tiny ($\pm$3.7)} & 58.9{\tiny ($\pm$2.2)} & 49.3{\tiny ($\pm$0.3)} & 82.4{\tiny ($\pm$0.5)} & 20.6{\tiny ($\pm$0.4)} & 28.5{\tiny ($\pm$2.6)}
  \\
  
  \textbf{AP} 
  & \underline{53.7} & \underline{48.8}{\tiny ($\pm$4.5)} & \underline{23.5}{\tiny ($\pm$0.8)} & \underline{80.6}{\tiny ($\pm$0.5)} & \underline{52.5}{\tiny ($\pm$0.4)} & \underline{85.6}{\tiny ($\pm$0.2)} & \underline{26.8}{\tiny ($\pm$0.6)} & \underline{58.1}{\tiny ($\pm$0.6)}
  \\
  
  \midrule
  \multicolumn{9}{c}{\textbf{Qwen3.5-0.8B}} \\
  \midrule

  Zero-shot 
  & 35.8 & 9.7{\tiny ($\pm$0.6)} & 13.3{\tiny ($\pm$1.4)} & 62.4{\tiny ($\pm$0.5)} & 42.4{\tiny ($\pm$0.2)} & 77.6{\tiny ($\pm$0.4)} & 16.7{\tiny ($\pm$0.7)} & 28.9{\tiny ($\pm$0.1)}
  \\
  
  LoRA 
  & 36.6 & 9.1{\tiny ($\pm$0.8)} & 14.1{\tiny ($\pm$1.7)} & 64.8{\tiny ($\pm$0.6)} & 43.8{\tiny ($\pm$0.2)} & 78.5{\tiny ($\pm$0.3)} & 16.5{\tiny ($\pm$3.5)} & 29.3{\tiny ($\pm$0.5)} 
  \\

  \midrule
  
  Many-shot (1K) 
  & \textbf{45.1} & \textbf{33.3}{\tiny ($\pm$2.5)} & \underline{17.3}{\tiny ($\pm$0.8)} & \textbf{73.1}{\tiny ($\pm$0.9)} & \textbf{47.4}{\tiny ($\pm$0.5)} & \textbf{80.5}{\tiny ($\pm$0.7)} & 13.8{\tiny ($\pm$1.3)} & \textbf{50.1}{\tiny ($\pm$0.9)}
  \\

  SITE 
  & 40.6 & 19.4{\tiny ($\pm$2.3)} & 15.3{\tiny ($\pm$1.5)} & 67.3{\tiny ($\pm$0.5)} & 45.6{\tiny ($\pm$0.3)} & 79.0{\tiny ($\pm$0.4)} & \underline{17.6}{\tiny ($\pm$0.4)} & 39.9{\tiny ($\pm$0.4)}
  \\
  
  ICV 
  & 37.7 & 19.2{\tiny ($\pm$7.1)} & 13.9{\tiny ($\pm$1.3)} & 63.0{\tiny ($\pm$0.5)} & 43.6{\tiny ($\pm$0.3)} & 78.0{\tiny ($\pm$0.6)} & 17.4{\tiny ($\pm$0.2)} & 28.5{\tiny ($\pm$0.3)}
  \\
  
  LTV 
  & 30.7 & 4.8{\tiny ($\pm$0.5)} & 10.3{\tiny ($\pm$2.2)} & 54.0{\tiny ($\pm$2.4)} & 41.8{\tiny ($\pm$0.7)} & 75.0{\tiny ($\pm$1.2)} & 15.8{\tiny ($\pm$0.8)} & 13.3{\tiny ($\pm$4.0)}
  \\
  
  \textbf{AP} 
  & \underline{44.3} & \underline{29.3}{\tiny ($\pm$2.3)} & \textbf{17.5}{\tiny ($\pm$1.5)} & \underline{71.5}{\tiny ($\pm$1.9)} & \underline{46.6}{\tiny ($\pm$0.6)} & \underline{79.8}{\tiny ($\pm$0.4)} & \textbf{18.5}{\tiny ($\pm$0.2)} & \underline{47.2}{\tiny ($\pm$1.6)}
  \\

  \midrule
  
  Many-shot (2K) 
  & \textbf{45.9} & \textbf{35.3}{\tiny ($\pm$1.8)} & \textbf{17.9}{\tiny ($\pm$0.7)} & \textbf{74.4}{\tiny ($\pm$0.9)} & \textbf{47.7}{\tiny ($\pm$0.1)} & \textbf{80.9}{\tiny ($\pm$0.3)} & 16.1{\tiny ($\pm$1.3)} & \textbf{49.2}{\tiny ($\pm$1.2)}
  \\
  
  SITE 
  & 40.5 & 18.4{\tiny ($\pm$2.2)} & 15.2{\tiny ($\pm$0.9)} & 67.8{\tiny ($\pm$1.0)} & 45.8{\tiny ($\pm$0.5)} & 79.1{\tiny ($\pm$0.5)} & \underline{17.8}{\tiny ($\pm$0.6)} & 39.5{\tiny ($\pm$0.4)}
  \\
  
  ICV 
  & 37.4 & 16.8{\tiny ($\pm$1.4)} & 14.1{\tiny ($\pm$1.5)} & 63.4{\tiny ($\pm$0.3)} & 43.6{\tiny ($\pm$0.4)} & 77.9{\tiny ($\pm$0.4)} & 17.3{\tiny ($\pm$0.4)} & 28.5{\tiny ($\pm$0.4)}
  \\
  
  LTV 
  & 29.7 & 6.4{\tiny ($\pm$1.2)} & 9.0{\tiny ($\pm$0.7)} & 51.8{\tiny ($\pm$0.8)} & 41.5{\tiny ($\pm$0.6)} & 74.8{\tiny ($\pm$1.1)} & 15.0{\tiny ($\pm$0.8)} & 9.4{\tiny ($\pm$1.3)}
  \\
  
  \textbf{AP} 
  & \underline{44.6} & \underline{32.7}{\tiny ($\pm$1.3)} & \underline{17.3}{\tiny ($\pm$1.7)} & \underline{71.9}{\tiny ($\pm$1.3)} & \underline{47.0}{\tiny ($\pm$0.9)} & \underline{80.0}{\tiny ($\pm$0.6)} & \textbf{18.4}{\tiny ($\pm$0.6)} & \underline{44.9}{\tiny ($\pm$0.2)} 
  \\

  \midrule
  
  Many-shot (4K) 
  & \textbf{46.5} & \textbf{33.4}{\tiny ($\pm$1.8)} & \underline{18.3}{\tiny ($\pm$1.3)} & \textbf{75.5}{\tiny ($\pm$0.5)} & \textbf{48.0}{\tiny ($\pm$0.3)} & \textbf{81.4}{\tiny ($\pm$0.2)} & \underline{18.2}{\tiny ($\pm$1.7)} & \textbf{50.5}{\tiny ($\pm$1.3)}
  \\
  
  SITE & 40.4 & 18.9{\tiny ($\pm$1.4)} & 15.0{\tiny ($\pm$1.4)} & 67.6{\tiny ($\pm$0.9)} & 45.6{\tiny ($\pm$0.5)} & 79.0{\tiny ($\pm$0.4)} & 17.7{\tiny ($\pm$0.8)} & 39.0{\tiny ($\pm$0.7)}
  \\
  
  ICV 
  & 37.2 & 16.3{\tiny ($\pm$1.5)} & 14.0{\tiny ($\pm$1.5)} & 63.5{\tiny ($\pm$0.2)} & 43.4{\tiny ($\pm$0.2)} & 77.8{\tiny ($\pm$0.4)} & 17.3{\tiny ($\pm$0.5)} & 28.5{\tiny ($\pm$0.0)} 
  \\

  LTV 
  & 29.0 & 4.1{\tiny ($\pm$1.6)} & 8.8{\tiny ($\pm$1.6)} & 49.5{\tiny ($\pm$1.1)} & 42.0{\tiny ($\pm$1.0)} & 75.3{\tiny ($\pm$1.2)} & 15.0{\tiny ($\pm$0.4)} & 8.0{\tiny ($\pm$0.3)}
  \\
  
  \textbf{AP} 
  & \underline{44.5} & \underline{28.2}{\tiny ($\pm$0.9)} & \textbf{18.5}{\tiny ($\pm$1.1)} & \underline{73.3}{\tiny ($\pm$1.3)} & \underline{46.8}{\tiny ($\pm$0.8)} & \underline{79.9}{\tiny ($\pm$0.5)} & \textbf{18.9}{\tiny ($\pm$0.1)} & \underline{45.9}{\tiny ($\pm$0.8)}
  \\

  \midrule
  
  Many-shot (8K) 
  & \textbf{46.5} & \textbf{34.7}{\tiny ($\pm$0.4)} & \textbf{19.1}{\tiny ($\pm$1.1)} & \textbf{76.1}{\tiny ($\pm$0.3)} & \textbf{47.9}{\tiny ($\pm$0.4)} & \textbf{81.5}{\tiny ($\pm$0.1)} & \underline{17.8}{\tiny ($\pm$2.8)} & \textbf{48.7}{\tiny ($\pm$0.8)}
  \\
  
  SITE 
  & 40.6 & 19.3{\tiny ($\pm$3.4)} & 15.0{\tiny ($\pm$1.1)} & 68.0{\tiny ($\pm$0.3)} & 45.9{\tiny ($\pm$0.2)} & 79.3{\tiny ($\pm$0.3)} & 17.7{\tiny ($\pm$0.5)} & 39.1{\tiny ($\pm$0.7)}
  \\
  
  ICV 
  & 37.3 & 16.9{\tiny ($\pm$2.1)} & 14.1{\tiny ($\pm$1.4)} & 63.5{\tiny ($\pm$0.7)} & 43.3{\tiny ($\pm$0.3)} & 77.9{\tiny ($\pm$0.5)} & 17.4{\tiny ($\pm$0.3)} & 28.3{\tiny ($\pm$0.2)}
  \\
  
  LTV 
  & 27.9 & 4.2{\tiny ($\pm$0.5)} & 7.8{\tiny ($\pm$2.5)} & 48.0{\tiny ($\pm$5.1)} & 40.7{\tiny ($\pm$2.1)} & 74.3{\tiny ($\pm$2.3)} & 13.2{\tiny ($\pm$0.6)} & 7.5{\tiny ($\pm$0.4)}
  \\
  
  \textbf{AP} 
  & \underline{44.3} & \underline{28.3}{\tiny ($\pm$2.0)} & \underline{17.4}{\tiny ($\pm$1.2)} & \underline{72.8}{\tiny ($\pm$0.7)} & \underline{47.0}{\tiny ($\pm$0.8)} & \underline{80.2}{\tiny ($\pm$0.4)} & \textbf{19.4}{\tiny ($\pm$0.5)} & \underline{45.3}{\tiny ($\pm$1.4)} 
  \\

  \midrule
  
  Many-shot (16K) 
  & \textbf{46.2} & \textbf{31.5}{\tiny ($\pm$0.9)} & \underline{19.4}{\tiny ($\pm$1.1)} & \textbf{76.0}{\tiny ($\pm$0.6)} & \textbf{48.0}{\tiny ($\pm$0.2)} & \textbf{81.5}{\tiny ($\pm$0.1)} & \textbf{18.9}{\tiny ($\pm$1.8)} & \textbf{48.1}{\tiny ($\pm$0.6)}
  \\

  SITE 
  & 40.1 & 18.3{\tiny ($\pm$2.8)} & 15.0{\tiny ($\pm$1.1)} & 68.3{\tiny ($\pm$0.6)} & 44.8{\tiny ($\pm$1.2)} & 78.3{\tiny ($\pm$1.1)} & \underline{17.5}{\tiny ($\pm$0.2)} & 38.4{\tiny ($\pm$0.8)}
  \\
  
  ICV 
  & 37.3 & 16.6{\tiny ($\pm$3.0)} & 14.0{\tiny ($\pm$1.2)} & 63.4{\tiny ($\pm$0.5)} & 43.5{\tiny ($\pm$0.4)} & 77.9{\tiny ($\pm$0.5)} & 17.3{\tiny ($\pm$0.4)} & 28.7{\tiny ($\pm$0.1)}
  \\
  
  LTV 
  & 27.2 & 4.2{\tiny ($\pm$0.3)} & 7.2{\tiny ($\pm$2.3)} & 43.4{\tiny ($\pm$6.2)} & 40.9{\tiny ($\pm$1.1)} & 74.5{\tiny ($\pm$1.2)} & 12.9{\tiny ($\pm$0.5)} & 7.2{\tiny ($\pm$0.2)}
  \\
  
  \textbf{AP} 
  & \underline{43.4} & \underline{20.1}{\tiny ($\pm$3.4)} & \textbf{19.5}{\tiny ($\pm$0.6)} & \underline{74.5}{\tiny ($\pm$0.8)} & \underline{46.8}{\tiny ($\pm$0.5)} & \underline{80.3}{\tiny ($\pm$0.2)} & \textbf{18.9}{\tiny ($\pm$0.7)} & \underline{43.7}{\tiny ($\pm$2.4)}
  \\

  \bottomrule
\end{tabular}}
\end{table}

Table~\ref{append:tab:generation} presents additional multi-token generation results on Mistral-7B-Instruct-v0.3, Gemma-2-2b-it, and Qwen3.5-0.8B across mathematical reasoning, translation, summarization, and constrained text generation.

On the additional models, \ours\ generally outperforms the evaluated implicit ICL baselines across generation tasks and demonstration-context budgets. Averaged across the five context budgets on Mistral-7B-Instruct-v0.3, \ours\ achieves 51.8 accuracy on GSM8K, compared with 41.1 for SITE and 51.9 for explicit many-shot ICL. On Gemma-2-2B-it, \ours\ exceeds many-shot ICL on GSM8K while maintaining comparable performance on FLORES. On Qwen3.5-0.8B, \ours\ generally outperforms these baselines and achieves competitive performance relative to explicit many-shot ICL on some tasks. These results suggest that the learned patches remain effective across different models.

\subsection{Amortized Inference Efficiency}

To further evaluate the trade-off between calibration costs and inference efficiency, we conduct a break-even analysis comparing AP with PrefixKV Caching on GSM8K using Llama-3.2-3B-Instruct with a fixed generation length of 512 tokens. We measure wall-clock time across 500 queries under context budgets ranging from 1K to 16K tokens. The break-even point is defined as the number of queries after which AP achieves lower cumulative wall-clock time than explicit ICL.

\begin{figure}
    \centering
    \includegraphics[width=1\linewidth]{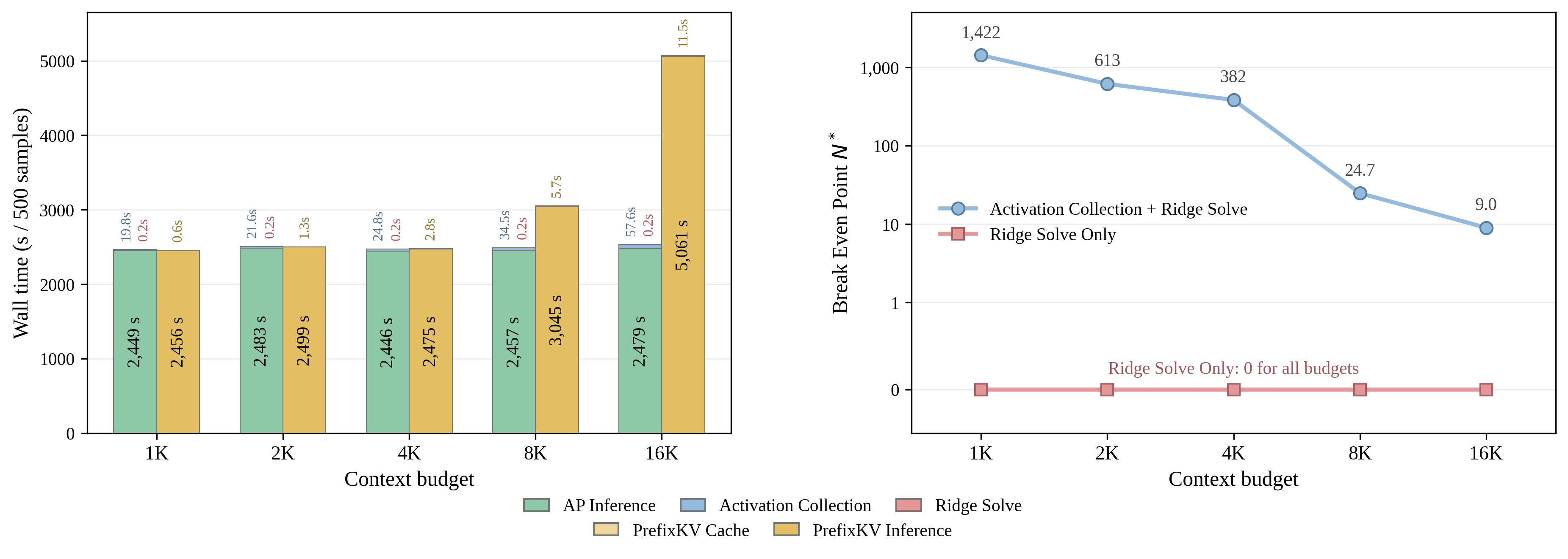}
    \caption{Break-even analysis of AP vs. PrefixKV Caching. Left: Wall-clock time for 500 queries across different context budgets. Right: Break-even points with and without activation collection costs.}
    \label{fig:break_even_point}
\end{figure}

As shown in Figure~\ref{fig:break_even_point}, when activation collection is excluded, the ridge regression step alone incurs negligible cost, making AP advantageous from the first query across all evaluated context budgets. When the full calibration cost, including activation collection and ridge regression, is considered, AP reaches the break-even point after approximately 25 queries at 8K and 9 queries at 16K. For shorter contexts (1K--4K), more queries are required to amortize the offline cost due to smaller per-query inference-time savings.

\section{Additional Ablation Studies}

\begin{figure}[t]
\centering
\includegraphics[width=1\linewidth]{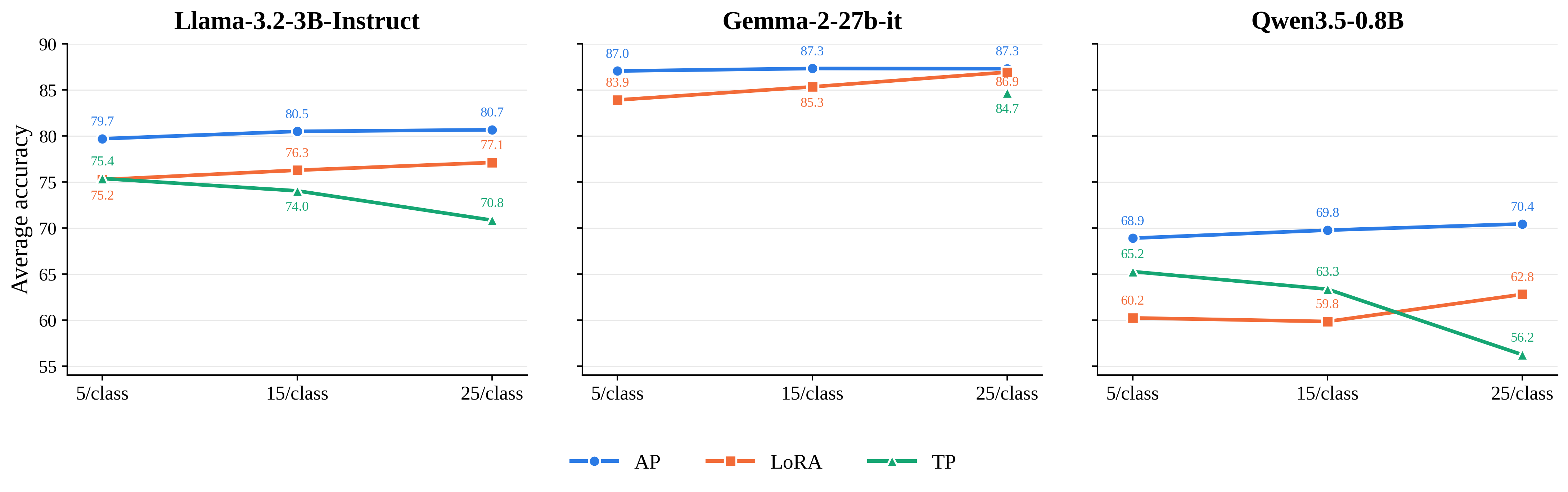}
\caption{Effect of calibration set size on Llama-3.2-3B-Instruct, Gemma-2-27b-it, and Qwen3.5-0.8B.
We report average accuracy across 12 classification benchmarks for LoRA, Thought Patching (TP), and \ours\ with 5, 15, and 25 examples per class.}
\label{fig:calibraion_num}
\end{figure}

\begin{figure}[!h]
\centering
\includegraphics[width=0.67\linewidth]{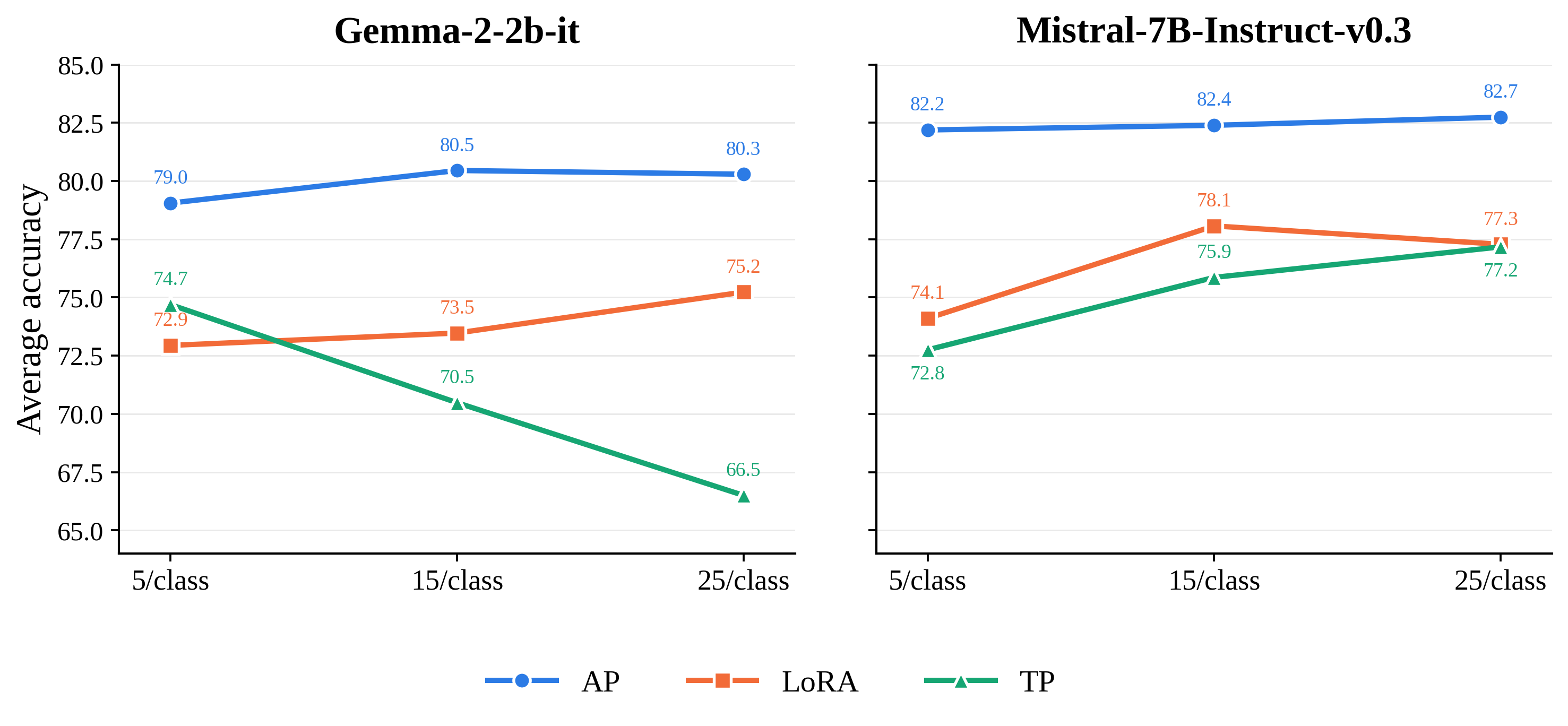}
\caption{Effect of calibration set size on Gemma-2-2b-it and Mistral-7B-Instruct-v0.3.
We report average accuracy across 12 classification benchmarks for LoRA, Thought Patching (TP), and \ours\ with 5, 15, and 25 examples per class.}
\label{fig:calibraion_num_append}
\end{figure}

\subsection{Effect of Calibration Set Size}

To examine the sensitivity of \ours\ to calibration set size, we compare its performance with LoRA and Thought Patching under different data budgets. We consider 5, 15 and 25 examples per class.


As shown in Figures~\ref{fig:calibraion_num} and~\ref{fig:calibraion_num_append}, \ours\ exhibits a clear advantage over LoRA and Thought Patching (TP) in the low-data regime. With only 5 examples per class, \ours\ achieves higher average accuracy than both baselines across all five evaluated models, while remaining competitive with the implicit ICL baselines reported in Tables~\ref{tab:classification} and~\ref{appendix:tab:classification}. This advantage over LoRA and TP is maintained as the calibration set size increases to 15 and 25 examples per class. These results suggest that \ours\ effectively captures task-specific behaviors induced by demonstrations using a small calibration set, without requiring backpropagation through the language model.

\subsection{Effect of Layer Selection}
To examine the contribution of different layer groups, we apply the learned patches to one quarter of the decoder layers at a time and compare their performance with patching all layers. We use the task-specific patch strength selected in the main experiments.



\begin{table*}[!h]
\caption{
Layer-group ablation of Attention-Patching on Llama-3.2-3B-Instruct across 12 classification benchmarks. We divide the decoder layers into four contiguous groups of equal size, where Q1 and Q4 denote the earliest and latest layer groups, respectively. ``All'' applies patches to every decoder layer.
}
\label{tab:abl_layer}
\centering
\small
\setlength{\tabcolsep}{2pt}
\begin{tabular}{@{}l@{\hspace{1pt}}|@{\hspace{1pt}}cccccccccccc@{\hspace{1pt}}|@{\hspace{1pt}}c@{}}
  \toprule

  Group
  & AG & SST-2 & TREC & CSQA & PIQA & SST-5
  & MR & MRPC & CB & COPA & CREAK & AI2SciE
  & Avg. \\
  
  \midrule
  None
  & 58.8 & 90.8 & 62.0
  & 76.6 & 75.6 & 33.8
  & 89.0 & 66.9 & 67.9
  & 94.0 & 79.2 & 87.0
  & 73.5 \\
  
  Q1
  & 66.8 & 91.9 & 67.2 & 76.1 & 75.8 & 31.3 & 89.6 & 66.5 & 61.8 & 92.8 & 69.8 & 86.9 & 73.0 \\

  Q2
  & 72.4 & 90.5 & 74.7 & 76.5 & 75.3 & 36.9 & 89.0 & 67.3 & 65.4 & 94.0 & 83.7 & 86.3 & 76.0 \\

  Q3
  & 63.3 & 89.8 & 61.5 & 76.5 & 75.6 & 34.5 & 89.2 & 67.6 & 62.5 & 93.6 & 78.6 & 87.3 & 73.3 \\

  Q4
  & 56.7 & 90.6 & 57.2 & 76.3 & 75.6 & 34.7 & 89.4 & 67.9 & 64.6 & 94.0 & 75.9 & 86.8 & 72.5 \\

  \textbf{All}
  & 85.8 & 93.0 & 83.9 & 76.2 & 75.8 & 46.7 & 91.1 & 68.7 & 79.6 & 93.2 & 86.0 & 87.7 & 80.7 \\
  \bottomrule
\end{tabular}
\end{table*}

Table~\ref{tab:abl_layer} evaluates the contribution of different layer groups by applying the learned patches to one quarter of the decoder layers at a time. Patching all layers achieves the highest accuracy of 80.7, outperforming every individual layer group. Based on this result, we apply the learned patches to all decoder layers by default.

\subsection{Sensitivity to Ridge Regularization}



\begin{table*}[!h]
\caption{
Effect of the ridge coefficient $\rho$ on Llama-3.2-3B-Instruct. We report average accuracy across 12 classification benchmarks.
}
\label{tab:abl_rho}
\centering
\small
\setlength{\tabcolsep}{2pt}
\begin{tabular}{@{}l@{\hspace{1pt}}|@{\hspace{1pt}}cccccccccccc@{\hspace{1pt}}|@{\hspace{1pt}}c@{}}
  \toprule

  \(\rho\)
  & AG & SST-2 & TREC & CSQA & PIQA & SST-5
  & MR & MRPC & CB & COPA & CREAK & AI2SciE
  & Avg. \\
  
  \midrule

  \textbf{1e-3}
  & 85.8 & 93.0 & 83.9 & 76.2 & 75.8 & 46.7 & 91.1 & 68.7 & 79.6 & 93.2 & 86.0 & 87.7 & 80.7 \\

  1e-2
  & 79.0 & 92.8 & 83.2 & 76.3 & 75.9 & 43.4 & 91.3 & 66.6 & 79.3 & 94.8 & 85.0 & 88.0 & 79.6 \\

  1e-1
  & 59.3 & 92.2 & 84.3 & 76.1 & 75.3 & 38.3 & 90.8 & 66.9 & 70.7 & 95.6 & 85.9 & 87.1 & 76.9 \\
  \bottomrule
\end{tabular}
\end{table*}

Table~\ref{tab:abl_rho} examines the sensitivity of \ours\ to the ridge coefficient $\rho$ on classification benchmarks. We vary $\rho$ from $1e-3$ to $1e-1$. Based on these results, we adopt $\rho = 1e-3$ as the default setting.

\section{Additional Representation Analysis}
\subsection{Layer-wise Structure on Additional Models}

\begin{figure}[!h]
\centering
\includegraphics[width=1\linewidth]{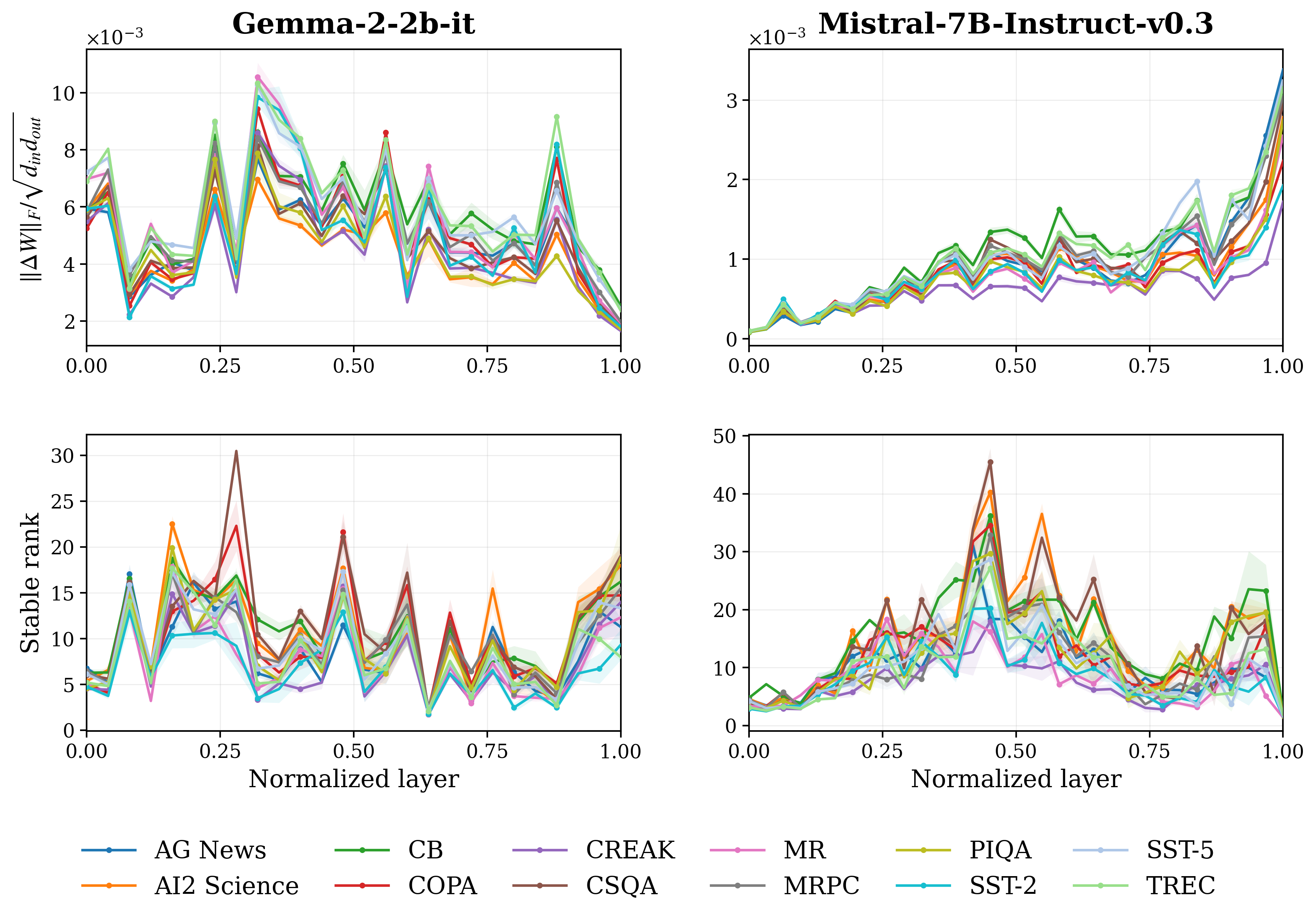}
\caption{Layer-wise structure of the learned attention patches across classification benchmarks. The top row shows the normalized Frobenius norm of the patch weights, while the bottom row shows their stable rank. Each column corresponds to a model architecture, and each curve represents one classification task.}
\label{fig:layer_patching_append}
\end{figure}



We provide additional representation analyses complementing the results presented in Section~\ref{sec:layer_wise_structure}.

Figure~\ref{fig:layer_patching_append} extends the layer-wise analysis of learned attention patches to additional model architectures, examining their magnitude and effective dimensionality through the normalized Frobenius norm and stable rank, respectively.

\subsection{NMSE Comparison}
\label{sec:mse_comparison}
\begin{figure}[!h]
\centering
\includegraphics[width=1\linewidth]{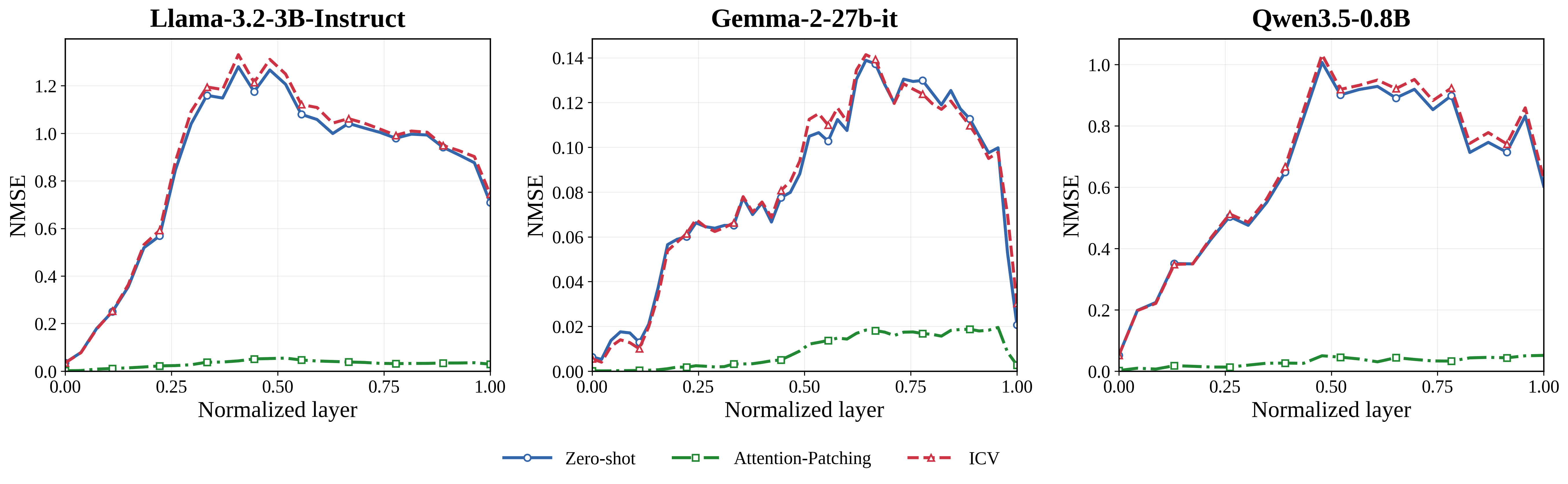}
\caption{Layer-wise NMSE of hidden states relative to many-shot ICL, averaged over 500 GSM8K samples and context lengths. Lower values indicate greater representational similarity.}
\label{fig:mse}
\end{figure}

\begin{figure}[!h]
\centering
\includegraphics[width=1\linewidth]{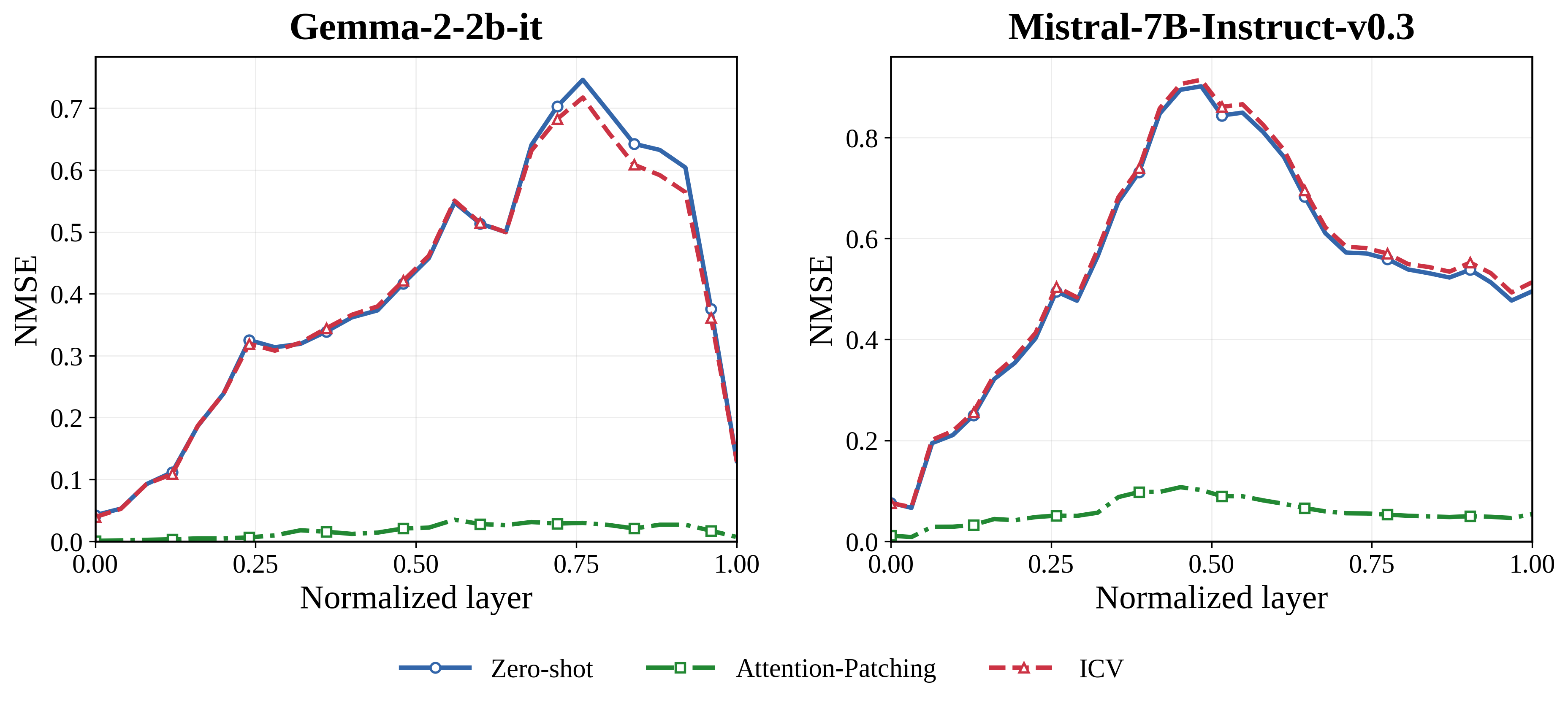}
\caption{Layer-wise NMSE of hidden states relative to many-shot ICL, averaged over 500 GSM8K samples and context lengths. Lower values indicate greater representational similarity.}
\label{fig:mse_append}
\end{figure}

Figures~\ref{fig:mse} and \ref{fig:mse_append} report the layer-wise normalized mean squared error (NMSE) between the layer-output hidden states produced by each method and those produced by explicit many-shot ICL. The results are evaluated on 500 randomly sampled GSM8K examples and averaged across the evaluated demonstration-context budgets for each model.
Across layers, \ours\ produces hidden states that are closer to those of explicit ICL than either zero-shot inference or ICV. 
The reduced representational distance to explicit ICL suggests that the learned patches recover aspects of its internal activation patterns, providing further evidence that \ours\ captures contextual adaptation beyond the final output predictions.

\end{document}